\documentclass[10pt,twocolumn,letterpaper]{article}

\usepackage{cvpr}              

\usepackage{graphicx}
\usepackage{amsmath}
\usepackage{amssymb}
\usepackage{booktabs}
\usepackage{multirow}
\usepackage{enumerate}
\usepackage{makecell}
\usepackage{colortbl}
\usepackage{rotating}

\newcommand\ChangeRT[1]{\noalign{\hrule height #1}}

\usepackage[pagebackref,breaklinks,colorlinks]{hyperref}

\definecolor{rel_c}{RGB}{83,155,101}
\definecolor{obj_c}{RGB}{199,96,73}
\definecolor{attr_c}{RGB}{73,124,194}
\definecolor{attr_c2}{RGB}{112,48,160}
\definecolor{act_c}{RGB}{180,138,38}

\usepackage[capitalize]{cleveref}
\crefname{section}{Sec.}{Secs.}
\Crefname{section}{Section}{Sections}
\Crefname{table}{Table}{Tables}
\crefname{table}{Tab.}{Tabs.}

\def\cvprPaperID{5317} 
\def\confName{CVPR}
\def\confYear{2023}

\definecolor{mygray}{gray}{0.87}

\begin{document}

\title{ANetQA: A Large-scale Benchmark for Fine-grained Compositional Reasoning over Untrimmed Videos}

\author{
	Zhou Yu\textsuperscript{\rm 1}\quad
	Lixiang Zheng\textsuperscript{\rm 1}\quad
	Zhou Zhao\textsuperscript{\rm 2}\quad
	Fei Wu\textsuperscript{\rm 2}\quad
	Jianping Fan\textsuperscript{\rm 1, 3}\quad
	Kui Ren\textsuperscript{\rm 4}\quad
	Jun Yu\textsuperscript{\rm 1}\thanks{Jun Yu is the corresponding author}\quad\quad
	\\
	\normalsize\textsuperscript{\rm 1} School of Computer Science, Hangzhou Dianzi University, China.\\
	\normalsize \textsuperscript{\rm 2}Colledge of Computer Science and Technology, Zhejiang University, China\\
	\normalsize \textsuperscript{\rm 3}AI Lab at Lenovo Research, China\\
	\normalsize \textsuperscript{\rm 4}School of Cyber Science and Technology, Zhejiang University, China\\
	{\fontfamily{pcr}\selectfont \small \{yuz, lxzheng, yujun\}@hdu.edu.cn,~\{zhaozhou, wufei, kuiren\}@zju.edu.cn,~jfan1@lenovo.com}\\
}

\maketitle

\begin{abstract}
   Building benchmarks to systemically analyze different capabilities of video question answering (VideoQA) models is challenging yet crucial. Existing benchmarks often use non-compositional simple questions and suffer from language biases, making it difficult to diagnose model weaknesses incisively. A recent benchmark AGQA \cite{grunde2021agqa} poses a promising paradigm to generate QA pairs automatically from pre-annotated scene graphs, enabling it to measure diverse reasoning abilities with granular control. However, its questions have limitations in reasoning about the \textbf{fine-grained} semantics in videos as such information is absent in its scene graphs. To this end, we present ANetQA, a large-scale benchmark that supports {fine-grained} compositional reasoning over the challenging untrimmed videos from ActivityNet \cite{caba2015activitynet}. Similar to AGQA, the QA pairs in ANetQA are automatically generated from annotated video scene graphs. The fine-grained properties of ANetQA are reflected in the following: (i) untrimmed videos with fine-grained semantics; (ii) spatio-temporal scene graphs with fine-grained taxonomies; and (iii) diverse questions generated from fine-grained templates. ANetQA attains \textbf{1.4 billion} unbalanced and \textbf{13.4 million} balanced QA pairs, which is an order of magnitude larger than AGQA with a similar number of videos. Comprehensive experiments are performed for state-of-the-art methods. The best model achieves 44.5\% accuracy while human performance tops out at 84.5\%, leaving sufficient room for improvement. 
\end{abstract}

\section{Introduction}\label{sec:intro}

\begin{figure}
	\begin{center}
		\includegraphics[width=\columnwidth]{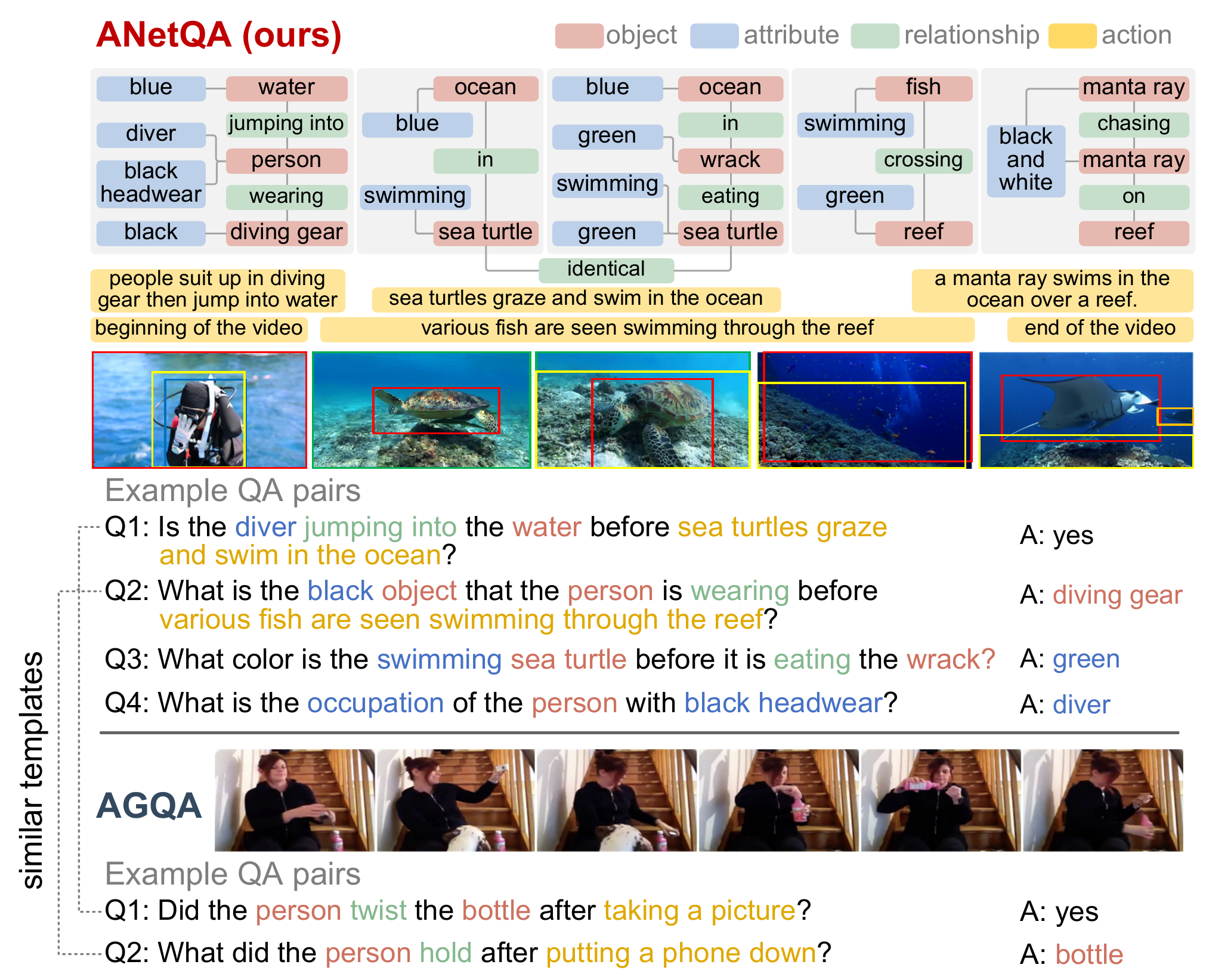}
		\vspace{-15pt}
		\caption{Comparisons of ANetQA and AGQA \cite{grunde2021agqa}. The QA pairs in both benchmarks are automatically generated from spatio-temporal scene graphs by using handcrafted question templates. Benefiting from the untrimmed long videos and fine-grained scene graphs, our questions require more fine-grained reasoning abilities than those in AGQA when similar templates are applied. Moreover, the newly introduced attribute annotations allow us to design many fine-grained question templates that are not supported in AGQA (\emph{e.g.}, ``\emph{what color}'' and ``\emph{what is the occupation}'').
		}
		\vspace{-25pt}
		\label{fig:agqa_example}
	\end{center}
\end{figure}

Recent advances in deep learning have enabled machines to tackle complicated video-language tasks that involve both video and language clues, \emph{e.g.}, video-text retrieval, video captioning, video temporal grounding, and video question answering. Among these tasks, video question answering (VideoQA) is one of the most challenging tasks as it verifies multiple skills simultaneously. Taking the question ``\emph{What is the black object that the person is wearing before various fish are seen swimming through the reef?}'' in \mbox{Figure \ref{fig:agqa_example}} as an example, it requires a synergistic understanding of both the video and question, together with spatio-temporal reasoning to predict an accurate answer.

To comprehensively evaluate the capabilities of existing VideoQA models, several prominent benchmarks have been established \cite{tapaswi2016movieqa,yu2019activitynet,lei2018tvqa,xu2017video,mun2017marioqa,yi2019clevrer,jang2017tgif}. Despite their usefulness, they also have distinct shortcomings. Some benchmarks use simulated environments to synthesize video contents \cite{mun2017marioqa,yi2019clevrer}, which provides controllable diagnostics over different reasoning skills. However, the synthetic videos lack visual diversity and the learned models on the benchmarks cannot generalize to real-world scenarios directly. Some real-world benchmarks generate QA pairs from off-the-shelf video captions \cite{zeng2017leveraging,xu2017video} or human annotations \cite{yu2019activitynet,lei2018tvqa,tapaswi2016movieqa,jang2017tgif}, which suffer from simple question expressions and biased answer distributions. These weaknesses may be exploited by models to make educated guesses to obtain the correct answers without seeing video contents \cite{yang2020gives,li2019repair}. 

One recent VideoQA benchmark AGQA poses a promising paradigm to address the above limitations \cite{grunde2021agqa}. AGQA is built upon the real-world videos from Charades \cite{sigurdsson2016hollywood}. In contrast to previous benchmarks, AGQA adopts a two-stage paradigm instead. For each video, a spatio-temporal scene graph over representative frames is first annotated by humans, which consists of spatially-grounded object-relationship triplets and temporally-grounded actions. After that, different types of questions are generated on top of the scene graph using corresponding question templates, enabling it to measure various reasoning abilities with granular control. Despite the comprehensiveness of AGQA, we argue that its foundation---the spatio-temporal scene graph---has limitations in representing the \emph{fine-grained} semantics of videos. Specifically, their scene graphs encode objects and relationships from limited taxonomies, which are \emph{not} fine-grained enough for generating questions that require reasoning about the detailed video semantics.

To this end, we introduce ANetQA\footnote{Note that there is a VideoQA benchmark ActivityNet-QA \cite{yu2019activitynet} whose QA pairs are fully annotated by humans. To avoid confusion, we name our benchmark ANetQA.}, a new benchmark that supports {fine-grained} compositional reasoning over complex web videos from ActivityNet \cite{caba2015activitynet}. Similar to the strategy of AGQA, the QA pairs in ANetQA are automatically generated from pre-annotated scene graphs. As shown in Figure \ref{fig:agqa_example}, we claim that ANetQA is more fine-grained than AGQA in terms of the following: 
\vspace{-5pt}
\begin{enumerate}[(i)]
	\item The benchmark is built upon untrimmed long videos with fine-grained semantics. Each video may involves multiple indoor or outdoor scenarios, containing complicated interactions between persons and objects.
	\vspace{-5pt} 
	\item The spatio-temporal scene graph consists of fine-grained objects (\emph{e.g.}, ``\emph{manta ray}'', ``\emph{diving gear}''), relationships (\emph{e.g.}, ``\emph{jumping into}'', ``\emph{chasing}''), attributes (\emph{e.g.}, ``\emph{swimming}'', ``\emph{black and white}''), and actions in natural language (\emph{e.g.}, ``\emph{a manta ray swims in the ocean over a reef}'').
	\vspace{-5pt} 
	\item Benefiting from the fine-grained scene graphs, we are able to design diverse question templates that requires fine-grained compositional reasoning (\emph{e.g.}, ``\emph{what color ...}'' and ``\emph{what is the occupation ...}''). 
\end{enumerate}
\vspace{-5pt} 

Benefiting from the above fine-grained characteristics, ANetQA obtains 1.4B unbalanced and 13.4M balanced QA pairs. To the best of our knowledge, ANetQA is the largest VideoQA benchmark in terms of the number of questions. Compared with the previous largest benchmark AGQA, ANetQA is an order of magnitude larger than it with a similar number of videos. We conduct comprehensive experiments and intensive analyses on ANetQA for the state-of-the-art VideoQA models, including HCRN \cite{le2020hierarchical}, ClipBERT \cite{lei2021less}, and All-in-One \cite{wang2022all}. The best model delivers 44.5\% accuracy while human performance tops out at 84.5\%, showing sufficient room for future improvement. 
The benchmark is available at here\footnote{\url{https://milvlg.github.io/anetqa}}.

\section{Related Work}
We briefly review the field of VideoQA in terms of methods and benchmarks. Since ANetQA is built upon ActivityNet \cite{caba2015activitynet}, we introduce ActivityNet and its derived benchmarks in particular.

\noindent\textbf{VideoQA approaches.} The research of visual question answering lies mainly in the image domain. A number of image question answering (ImageQA) methods have been developed to push state-of-the-art performance on public benchmarks successively \cite{fukui2016multimodal,yu2018beyond,yu2019deep,jiang2020defense}. As a natural extension of the ImageQA task, VideoQA is more challenging as it requires effective temporal representation modeling and spatio-temporal reasoning. Existing studies explore end-to-end neural networks in conjunction with hierarchical representations \cite{zhao2017video,xu2017video}, memory networks \cite{tapaswi2016movieqa,na2017read,gao2018motion}, and graph networks \cite{huang2020location,liu2021hair,xiao2022video}. 
\begin{table*}
	\small
	\begin{tabular}{l|ccc|cc|cccc}
		& \multicolumn{3}{c|}{video} & \multicolumn{2}{c|}{question} & \multicolumn{4}{c}{grounding taxonomy} \\
		& type &\#videos & avg. len. &\#QA pairs & \#templates & \#objects & \#relations & \#attributes & \#actions \\
		\ChangeRT{1.3pt}
		CLEVRER\cite{yi2019clevrer}  &synth.& 20K & 5s &305K & 5 & 1 & 2 & 13 & 3 \\
		TVQA+\cite{lei2019tvqa+} & real & 4.2K & 7.2s & 29.4K & - & \textbf{2,527} & - & - & open\\
		HowtoVQA69M\cite{yang2021just}  &real&  \textbf{69M} & 12.1s & 69M & - & - & - & - & open\\
		AGQA\cite{grunde2021agqa}  &real& 9.6K &30s& 192M/3.9M & 28 & 36 & 44 & - & 157\\
		\hline
		\hline
		\textbf{ANetQA} &real & 11.5K & \textbf{180s} & \textbf{1.4B/13.4M} & \textbf{119} & 2,072 & \textbf{86} & \textbf{618} & open\\
	\end{tabular}
	\caption{\textbf{Comparisons of ANetQA and other representative large-scale VideoQA benchmarks}. 
		Benefiting from the fine-grained video and grounding annotations, ANetQA attains massive fine-grained questions and is an order of magnitude larger than the current largest benchmarks \cite{grunde2021agqa,yang2021just} in terms of the number of QA pairs. ``open'' indicates the grounded actions are depicted in natural language.}
	\vspace{-10pt}
\end{table*}
Motivated by the encouraging success of Transformers \cite{vaswani2017attention} in various NLP \cite{devlin2019bert,raffel2020exploring}, CV \cite{dosovitskiy2020image,liu2021swin}, and multimodal tasks \cite{lu2019vilbert,chen2020uniter,cui2021rosita}, Transformer-based approaches have become the mainstream of recent VideoQA research. Early approaches only exploit the Transformer architecture and train models from scratch \cite{li2019beyond,jin2019multi}. More recently, pretrained Transformer models on large-scale datasets have shown effectiveness when finetuned on VideoQA tasks. Some approaches incorporate the pretrained language Transformers \cite{khan2020mmft,yang2020bert} or multimodal Transformers on image-text pairs \cite{lei2021less} to improve VideoQA performance. Some other studies perform video-language pretraining directly on massive video-text pairs, which learn better multimodal representations and achieve state-of-the-art performance on various VideoQA benchmarks \cite{yang2021just,zellers2021merlot,fu2021violet,wang2022all}.   
\vspace{5pt}
\\
\noindent\textbf{VideoQA benchmarks.} The rapid progress in VideoQA is inextricably related to the established benchmarks. Existing VideoQA benchmarks can be categorized into two groups based on whether their videos are synthesized by simulation \cite{mun2017marioqa,yi2019clevrer} or collected from the real world \cite{tapaswi2016movieqa,lei2018tvqa,zadeh2019social,yu2019activitynet,xiao2021next,grunde2021agqa,xu2017video,maharaj2017dataset,zeng2017leveraging,kim2017deepstory,yang2021just}. The synthesized benchmarks can easily obtain massive QA pairs without human annotations. Their synthetic nature also enables granular control over reasoning abilities and language biases. However, the synthesized videos are often short and lack visual diversity, making it difficult to generalize the learned models to real-world scenarios.  

Establishing VideoQA benchmarks on real-world videos requires human annotations inevitably. Early benchmarks rely on the associated video captions to generate QA pairs automatically \cite{maharaj2017dataset,xu2017video,zeng2017leveraging,zhu2017uncovering}. Although these captions are annotated by humans, they are often too general to cover all the fine-grained semantics in videos. This makes these benchmarks be dominated by simple questions that lack detailed information. To obtain fine-grained and diverse questions, some recent benchmarks have been established by asking annotators to design questions of specific reasoning abilities, \emph{e.g.}, object localization \cite{lei2019tvqa+}, relationship recognition \cite{yu2019activitynet}, and causality analysis \cite{xiao2021next}. Nevertheless, prohibitive annotation costs restrict the sizes of these benchmarks and free-form question expressions lead to severe language biases. One recent benchmark AGQA introduces a new paradigm to automatically generate QA pairs upon video scene graphs \cite{grunde2021agqa}. Through the composition of scene graph elements, AGQA is orders of magnitude larger than its counterparts.
Similar to AGQA, our ANetQA is also built upon spatio-temporal scene graphs. In contrast to AGQA, ANetQA shows its {fine-grained} characteristics in terms of the videos, annotated scene graphs, and generated questions. Detailed comparisons of ANetQA and other representative large-scale VideoQA benchmarks are shown in Table \ref{fig:agqa_example}.
\vspace{5pt}
\\
\noindent\textbf{ActivityNet and its derivatives.} ActivityNet (\emph{abbr.} ANet) is one of the most important video recognition benchmarks \cite{caba2015activitynet}. It consists of 20K untrimmed videos from 200 activity classes, including both indoor and outdoor scenarios. The benchmark is challenging as its videos contain rich semantics. Therefore, some derived benchmarks are built upon ANet to provide fine-grained annotations \cite{krishna2017dense,zhou2019grounded}. ANet-Captions \cite{krishna2017dense} annotates each video with multiple temporally-grounded captions. ANet-Entities \cite{zhou2019grounded} provides spatially-grounded bounding boxes for the noun phrases mentioned in the captions. We establish our ANetQA based on the annotations of these two benchmarks . 

\section{The ANetQA Benchmark}
ANetQA is a large-scale VideoQA benchmark to measure a variety of spatio-temporal reasoning abilities at a fine-grained level. In this section, we first provide an overview of the construction process of our benchmark and then introduce the key stages in detail.
 
\subsection{Overview}
The videos in ANetQA are derived from ActivityNet \cite{caba2015activitynet}. As mentioned above, we leverage the auxiliary annotations on ActivityNet \cite{zhou2019grounded,krishna2017dense} to reduce the annotation costs during the construction of our benchmark. These result in 11,525 videos in total, which are comprised of 9,155 and 2,370 videos in the \texttt{train} and \texttt{val} splits of ActivityNet, respectively. We keep the \texttt{train} split unchanged and further divide the \texttt{val} split evenly into a new \texttt{val} split of 1,185 videos and a \texttt{test} split of 1,185 videos.

Next, we annotate each video with a spatio-temporal scene graph via crowdsourcing. Each video has been annotated with temporal-grounded captions \cite{krishna2017dense} and spatially-grounded objects from a few representative frames \cite{zhou2019grounded}, For each frame, we first clean the mislabeled objects and complement the omitted objects, and then annotate each object with fine-grained relationships and attributes. The accomplished scene graph annotations consist of 118K objects, 83K relationships, 1M attributes, and 16K natural language actions across 43K representative video frames.

Finally, we handcraft a variety of templates to generate linguistically diverse QA pairs with both grammatical and logical guarantees. By composing the elements in the scene graphs and then filling them into proper template slots, we obtain 1.4B unbalanced and 13.4M balanced QA pairs. 

\captionsetup[subfigure]{font=normalsize}
\begin{figure*}
	\centering
	\begin{subfigure}[h]{0.225\linewidth}
		\includegraphics[width=\linewidth]{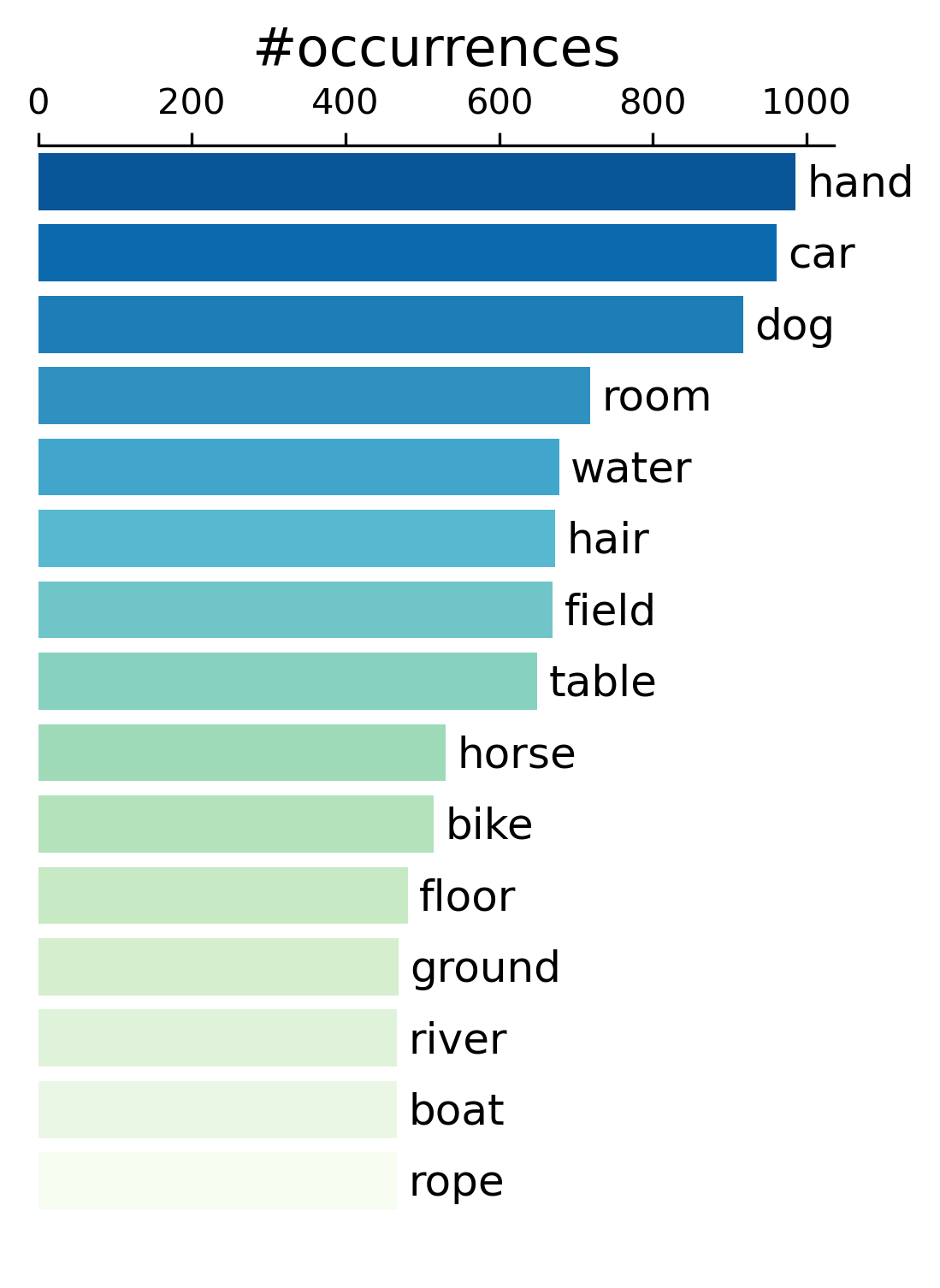}
		\caption{object distribution}\label{fig:obj_dist}
	\end{subfigure}
	\begin{subfigure}[h]{0.237\linewidth}
		\includegraphics[width=\linewidth]{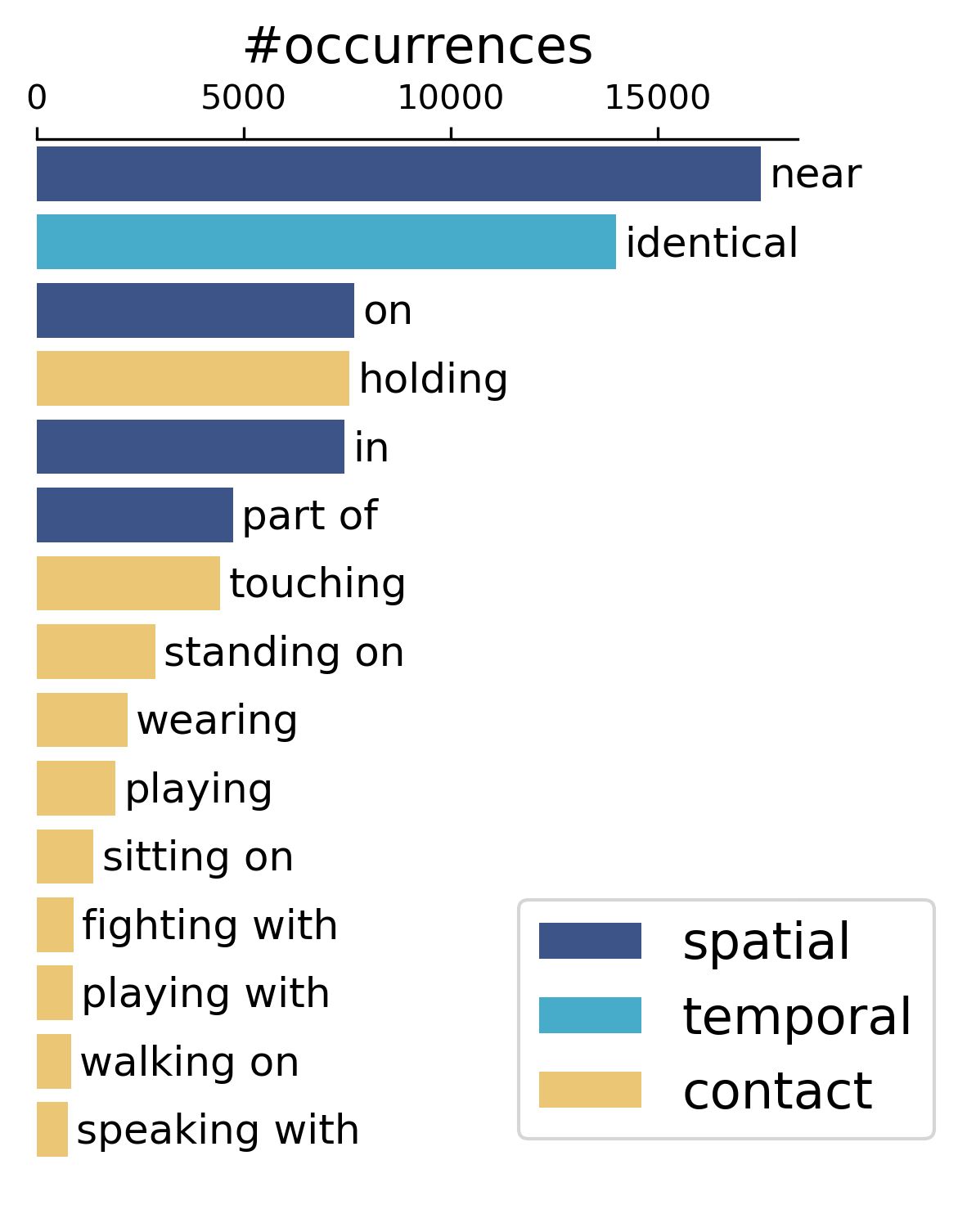}
		\caption{relationship distribution}\label{fig:rel_dist}
	\end{subfigure}
	\begin{subfigure}[h]{0.52\linewidth}
		\includegraphics[width=\linewidth]{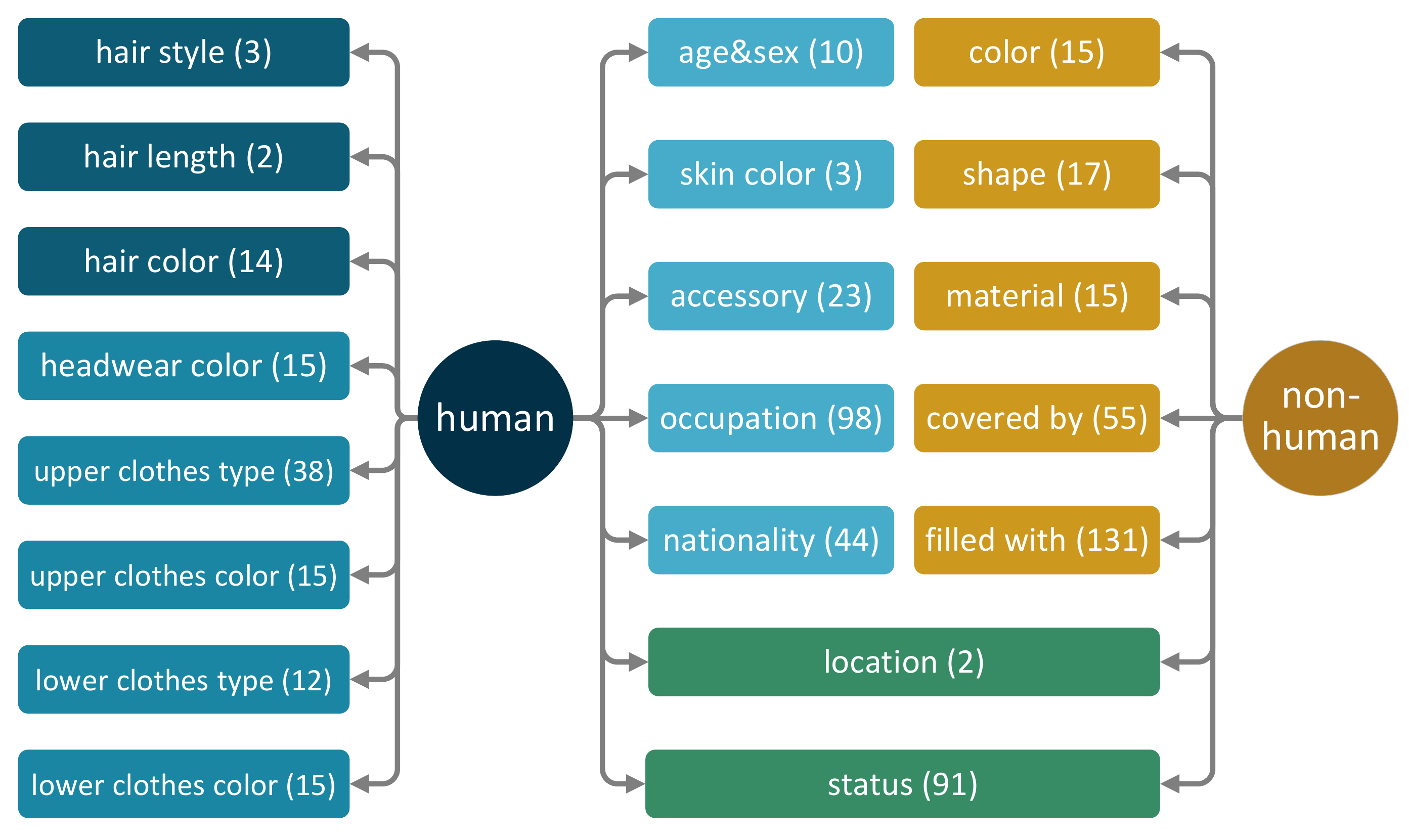}
		\caption{attribute hierarchy}\label{fig:attr_onto}
	\end{subfigure}
	\vspace{-5pt}
	\caption{\textbf{Statistics of the annotated video scene graphs}. We visualize the distributions of the top-15 (a) object occurrences and (b) relationship occurrences. The attributes form a hierarchical taxonomy shown in (c), where the values in the parentheses indicate the number of bottom-level attributes to be annotated. More details are provided in the supplementary material.}
	\vspace{-15pt}
	\label{fig:att_unit}
\end{figure*}

\subsection{Fine-grained Video Scene Graph Annotation}

\noindent\textbf{Representative frames.} Annotating a scene graph over all video frames is impractical. Similar to \cite{grunde2021agqa}, each of our scene graph is annotated over a few representative frames in a video. Concretely, we use the selected frames from ANet-Entities \cite{zhou2019grounded} as the initialization, which cover the key semantics of all the action segments in ANet-Captions \cite{krishna2017dense}. After that, we manually check and filter out those frames that hamper further annotation, \emph{i.e.}, the frames do not contain any meaningful objects or contain too many objects from the same class. Finally, we obtain 43K frames for further annotation, which indicates an average number of 3.69 frames per video\footnote{The number of sampled frames in our ANetQA is much lower than that of AGQA (3.69 \emph{vs.} 24.4 on average). The motivation derives from our observation that the scene graph elements barely change within an action segment. With a limited annotation budget, we favor the annotation \emph{density} in one frame rather than the annotation \emph{scale} across many frames.}.
\vspace{5pt}
\\
\noindent\textbf{Objects.} ANet-Entities also provides object-level annotations for all the selected frames. Each object is annotated with a bounding box and a noun phrase (\emph{e.g.}, ``\emph{a young woman}'', ``\emph{a black jacket}''). To better organize the object annotations, we first extract nouns from the noun phrases and convert them into a set of object labels. After that, we merge the synonymous object labels (\emph{e.g.}, ``\emph{mountain}'' and ``\emph{hill}'', ``\emph{saxophone}'' and ``\emph{sax}''). Finally, we ask annotators to go through all the selected frames to refine the annotations, including object augmentation, label correction, and bounding box calibration. By doing the above, we obtain a total number of 118K objects of 2,072 classes over the selected frames. The top most frequent classes are shown in Figure \ref{fig:obj_dist}. We exclude the most frequent class ``\emph{person}'' for better visualization.   
\vspace{5pt}
\\
\noindent\textbf{Relationships.} Beyond recognizing objects, predicting pairwise relationships between two objects is also important for scene understanding. Referring to the taxonomy in AGQA, we design a set of 86 relationships containing 81 contact relationships (\emph{e.g.}, ``\emph{holding}'', ``\emph{riding}'', ``\emph{wearing}''), 4 spatial relationships (``\emph{near}'', ``\emph{on}'', ``\emph{in}'', ``\emph{part of}'')\footnote{As the viewpoints of our videos are varied, we exclude two spatial relationships (``\emph{in front of}'' and ``\emph{behind}'') in AGQA to avoid ambiguity.}, and 1 temporal relationship (``\emph{identical}''). Our contact relationship categories are broader than AGQA (81 \emph{vs.} 16), because: (i) our videos contain both indoor and outdoor scenarios while AGQA only contains indoor ones; (ii) our relationships contain interactions between two arbitrary objects (\emph{i.e.}, human-object, human-human, and object-object interactions) while AGQA only contains human-object interactions. For each paired objects in one frame, annotators are asked to label \emph{at most} one spatial relationship and one contact relationship, respectively. The ``\emph{identical}'' temporal relationship indicates the objects in different frames refer to the same instance, which is used to provide indirect references of objects during question generation. Unlike other manually annotated relationships, this relationship is automatically obtained from the annotated attributes, which will be described below. The relationship occurrences follow a long-tail distribution and we illustrate the top most frequent classes in Figure \ref{fig:rel_dist}. 
\vspace{5pt}
\\
\noindent\textbf{Attributes.} To distinguish the fine-grained discrepancies between two objects, especially when they share the same object label, we need attribute annotations. 
Different from the single-label object taxonomy, the attribute taxonomy has a \emph{multi-label} nature in that each object has multiple attributes. Moreover, the attributes for different objects are different. To address the challenges above, we handcraft a \emph{hierarchical} attribute taxonomy by taking the characteristics of our annotated objects into consideration. As shown in Figure \ref{fig:attr_onto}, our attribute taxonomy includes three levels. At the top level, we categorize all the object classes into the \emph{human} and \emph{non-human} groups. For each group at the middle level, we design a set of representative attribute types (\emph{e.g.}, ``\emph{hair style}'' and ``\emph{skin color}`` for the \emph{human} group, ``\emph{shape}'' and ``\emph{material}'' for the \emph{non-human} group). A few attribute types like ``\emph{location}'' and ``\emph{status}'' are shared across the two groups. At the bottom level, we provide a set of attribute labels for each attribute type (\emph{e.g.}, ``\emph{long hair}'' and ``\emph{short hair}'' for the \emph{hair length} attribute type). For each object, annotators are asked to label the bottom-level attributes thoroughly. Due to space limitations, we only show the numbers of attributes at the bottom level in the figure. We have annotated 1M attributes over 118K objects, with an average number of 8.6 attributes per object. 

As a by-product, the annotated attributes can facilitate the annotation process of the ``\emph{identical}'' relationship. Specifically, if two objects in different frames have the same object label, we calculate their overlapping ratio of the annotated attributes. The pairs that surpass a confidence threshold are manually checked to ensure correctness. 

To the best of our knowledge, our benchmark is the \emph{first} attempt to provide large-scale and hierarchical attribute annotations for grounded objects in real-world videos.  
\vspace{5pt}
\\
\noindent\textbf{Actions.} In contrast to the objects, attributes, and relationships above, the action segments over specific time intervals of the video often contain much richer semantics. Using a simple label may lose the essential semantics of the action. Therefore, we use a natural language caption to describe each action segment in detail, which has been provided in ANet-Captions \cite{krishna2017dense}. However, some long captions are syntactically complex and are hard to be used for question generation. To this end, we set the maximum length of a caption to 10 and filter out those captions exceeding this threshold. This results in 16K temporally-grounded captions with an average length of 8.1 words. 

\subsection{Compositional QA Generation}
On top of the annotated spatio-temporal scene graphs, we aim to generate massive questions for diverse reasoning abilities. As shown in Table \ref{table:ques_stat}, we design a set of 21 question types to cover diverse reasoning skills in varying degrees of complexities. Each question type is categorized into one of the five structures (query, verify, choose, compare, and logic), which refers to the intention of the question. To fulfill the functionality of different question types, we handcraft at least one template for each question type, resulting in 119 grammatical and logical question templates. 
Similar to AGQA, we design a functional program for each template that traverses and composes the elements in the scene graphs, and fills them into proper template slots to produce compositional QA pairs automatically. 
\begin{table}
	\small
	\begin{tabular}{l|ccrr}
		type & structure & \#templ.\hspace{-1em} &\makecell{\#unbal.} &\makecell{\#bal.}\\
		\ChangeRT{1.3pt}
		attrRelWhat$^\dag$ & query & 30 & 169.5M & 2.63M \\
		attrWhat$^\dag$ & query & 15 & 70.4M & 1.43M \\
		relWhat & query & 1 &33.1M & 1.01M \\
		objRelWhere & query & 2 & 2.5M &0.55M \\
		objRelWhat & query & 2 & 7.1M & 0.56M \\
		objWhere & query & 1 & 2.9M & 0.43M \\
		objWhat & query & 1 & 0.5M & 0.14M \\
		objExist & verify & 1 & 51.7M & 1.00M \\
		objRelExist & verify & 1 & 98.3M & 0.94M \\
		actExist & verify &1 & 0.4M & 0.08M \\
		objRelWhatChoose & choose & 2 & 347.0M & 0.57M \\
		objWhatChoose & choose & 1 & 180.5M & 0.55M \\
		attrRelWhatChoose$^\dag$ & choose & 36 & 149.5M & 0.42M \\
		attrWhatChoose$^\dag$ & choose & 18 & 85.1M & 0.40M \\
		attrCompare$^\dag$ & compare & 1 & 138.0M & 2.02M \\
		attrSame$^\dag$ & compare & 1 & 0.09M & 0.01M \\
		actTime & compare & 1 & 0.01M & 0.01M \\
		actLongerVerify	 & compare & 1 & 0.01M & 0.01M \\
		actShorterVerify & compare & 1 & 0.01M & 0.01M \\
		andObjRelExist & logic & 1 & 20.2M & 0.35M \\
		xorObjRelExist & logic & 1 & 20.2M & 0.35M \\
		\hline
		\hline
		\textbf{\normalsize{overall}}	& - & \textbf{119} & \textbf{1.4B} & \textbf{13.4M}\\
	\end{tabular}
	\normalsize
	\vspace{-5pt}
	\caption{\textbf{Statistics of the generated questions}. Each question type belongs to a certain structure and contains at least one template. More details are provided in the supplementary material. $^\dag$: new question types that are not supported in AGQA.
	}\label{table:ques_stat}
	\vspace{-20pt}
\end{table}

Compared to the question types in AGQA, our major improvements lie in that we introduce 6 extra types with respect to attributes (\emph{i.e.}, the types starting with `attr' in Table \ref{table:ques_stat}). The annotated rich attributes enable us to design up to 101 question templates (\emph{e.g.}, ``\emph{what color is ...}'', ``\emph{what is the shape of ...}''), resulting in 612.6M unbalanced and 6.9M balanced QA pairs. Furthermore, the attribute annotations are also used to describe objects in almost all the rest templates (\emph{e.g.}, ``\emph{what is the relationship between the} \texttt{[attribute][object]} \emph{and} \texttt{[attribute][object]}\emph{?}''). The introduction of attributes not only provides a more precise description of the referred object but also increases the reasoning steps of the generated questions. It is worth noting that although we can describe an object in great detail (\emph{e.g.}, ``\emph{a walking young woman wearing green t-shirt and sunglasses}''), this would lead to a risk of combinational explosion and affect the readability of the questions. Therefore, we set the maximum number of attributes used in each question to two.
 
\begin{figure}
	\centering
	\begin{subfigure}[h]{0.49\columnwidth}
		\includegraphics[width=\linewidth]{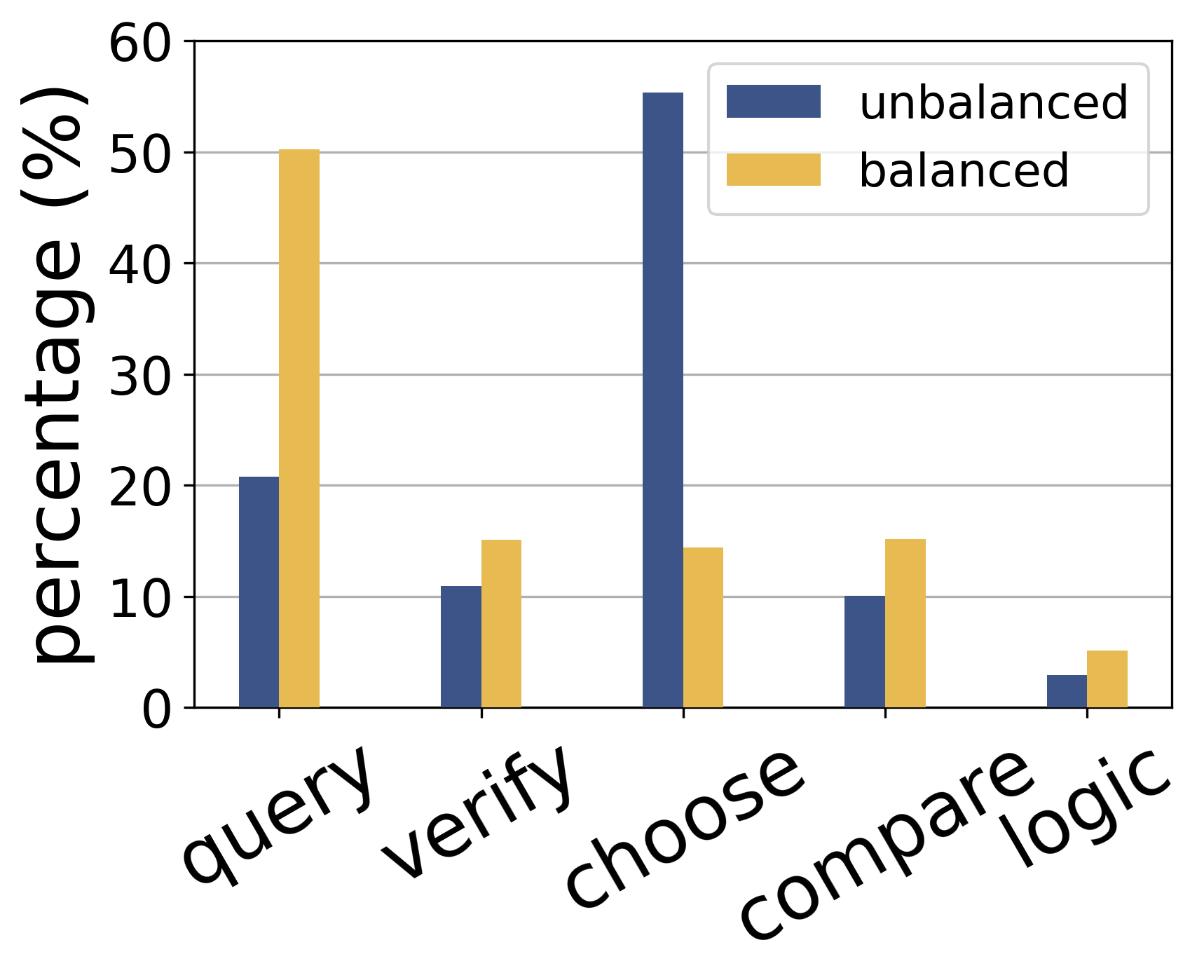}
		\caption{question balancing}\label{fig:ques_dist}
	\end{subfigure}
	\begin{subfigure}[h]{0.49\columnwidth}
		\includegraphics[width=\linewidth]{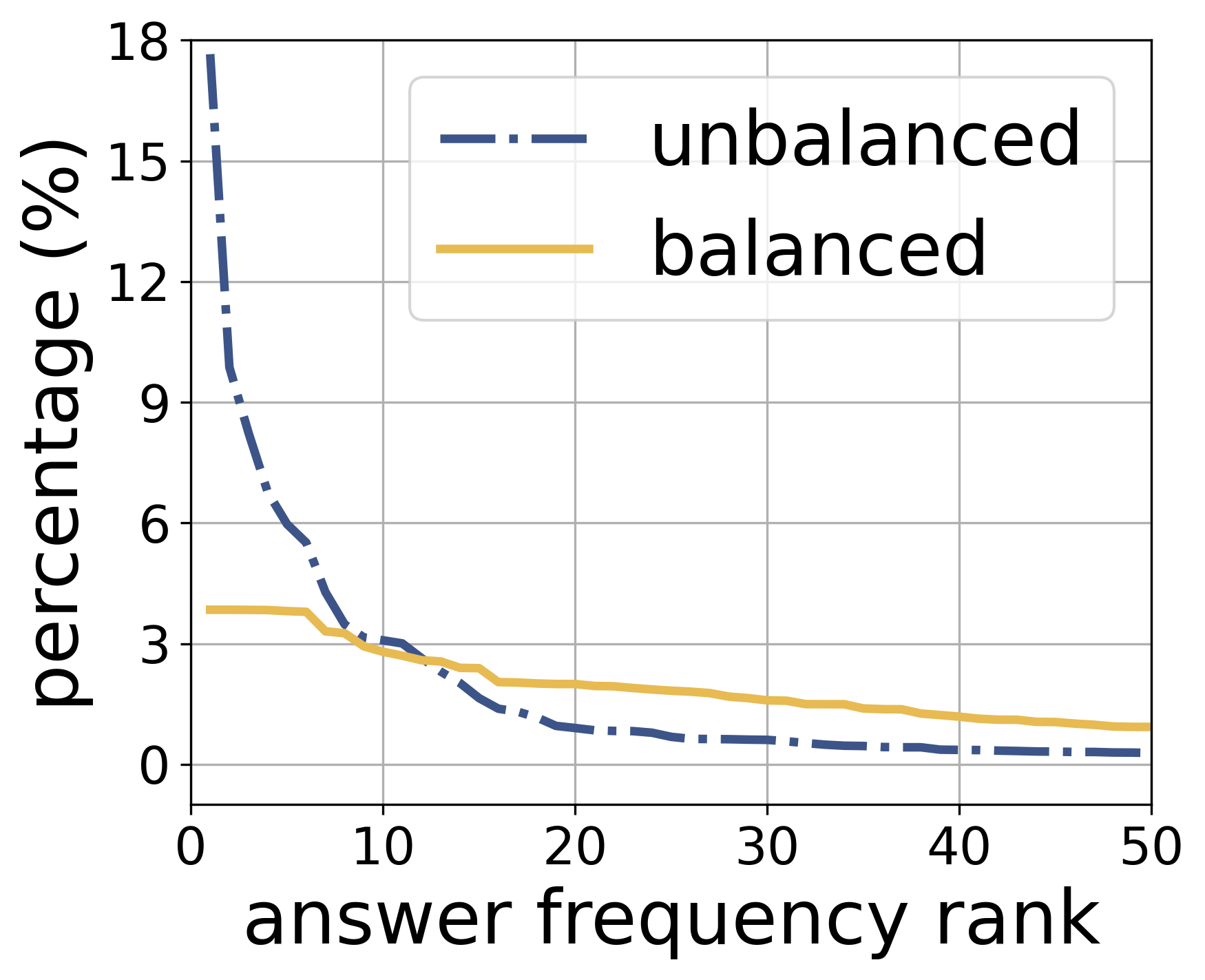}
		\caption{answer balancing}\label{fig:ans_dist}
	\end{subfigure}
	\vspace{-5pt}
	\caption{\textbf{Distributions before and after balancing.} (a) The question balancing is performed on question structures to adjust the percentages of the {query}/{verify}/{choose}/{compare}/{logic} questions to 50\%/15\%/15\%/15\%/5\%. (b) The answer balancing is conducted on each question template to make its answers follow a uniform distribution. Its effect to the global answer distribution can be observed from the change in the distributions of the top 50 most frequent open answers.}
	\label{fig:qa_dist_comp}
	\vspace{-10pt}
\end{figure}

Using the above question templates, we obtain 1.4 billion QA pairs. These QA pairs are \emph{unbalanced} and have strong language biases that models can exploit. We conduct composite balancing strategies on both the questions and answers. Following the question structure distribution in balanced AGQA, our question balancing strategy adjusts the percentages of the {query}/{verify}/{choose}/{compare}/{logic} questions to 50\%/15\%/15\%/15\%/5\%, as shown in Figure \ref{fig:ques_dist}. 
While maintaining these percentages above, we conduct answer balancing within each question template to make sure that its answers are uniformly distributed (unbiased). In Figure \ref{fig:ans_dist}, we visualize the global answer distributions of the unbalanced and balanced sets in terms of the top-50 most frequent open answers (\emph{i.e.}, the answers to the \emph{query} structure questions). The obtained results demonstrate the effectiveness of our balancing strategies.

\begin{figure}	
	\centering
	\begin{subfigure}[h]{0.49\columnwidth}
		\includegraphics[width=\linewidth]{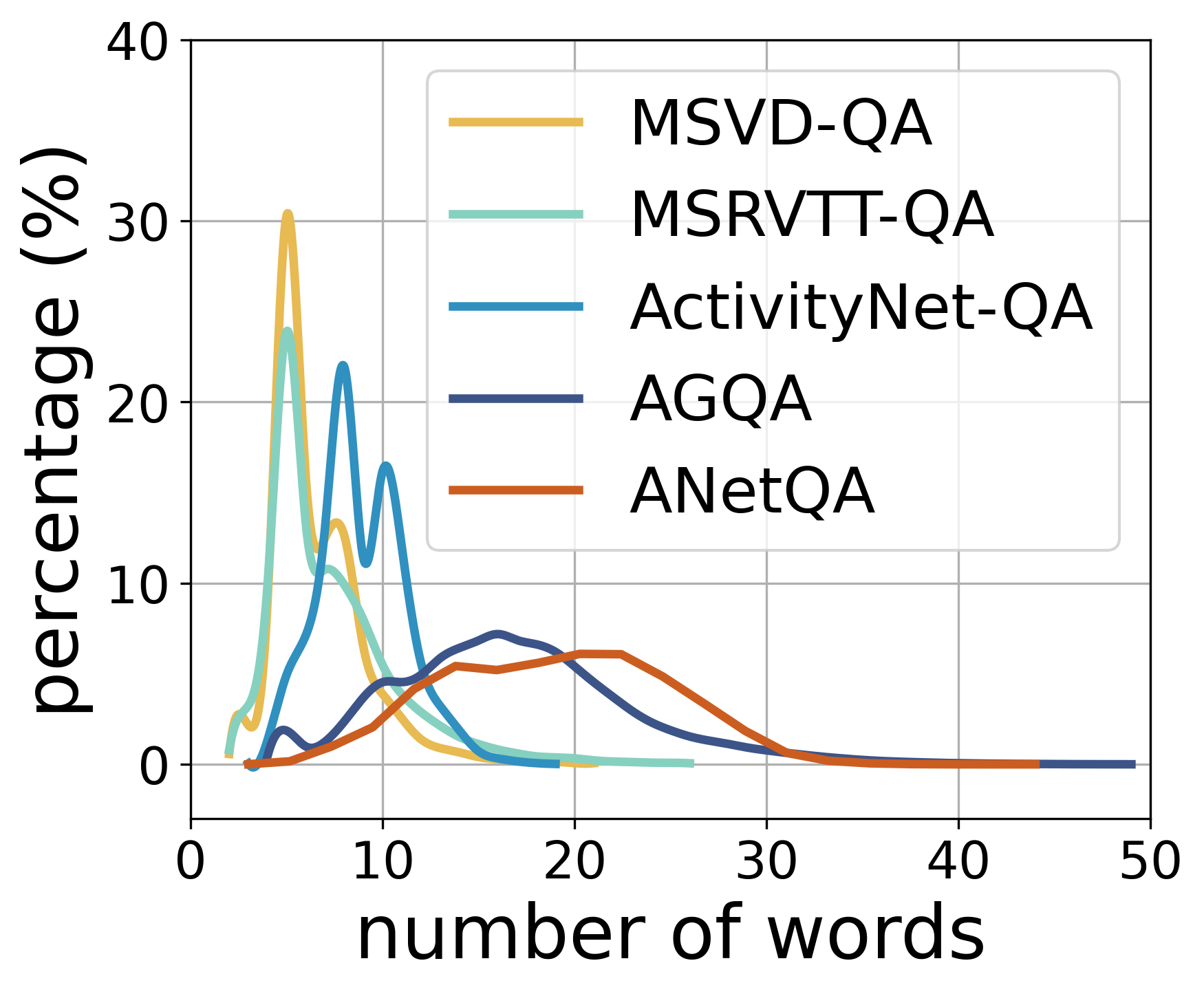}
		\caption{question lengths}\label{fig:qlen_dist}
	\end{subfigure}
	\begin{subfigure}[h]{0.49\columnwidth}
		\includegraphics[width=\linewidth]{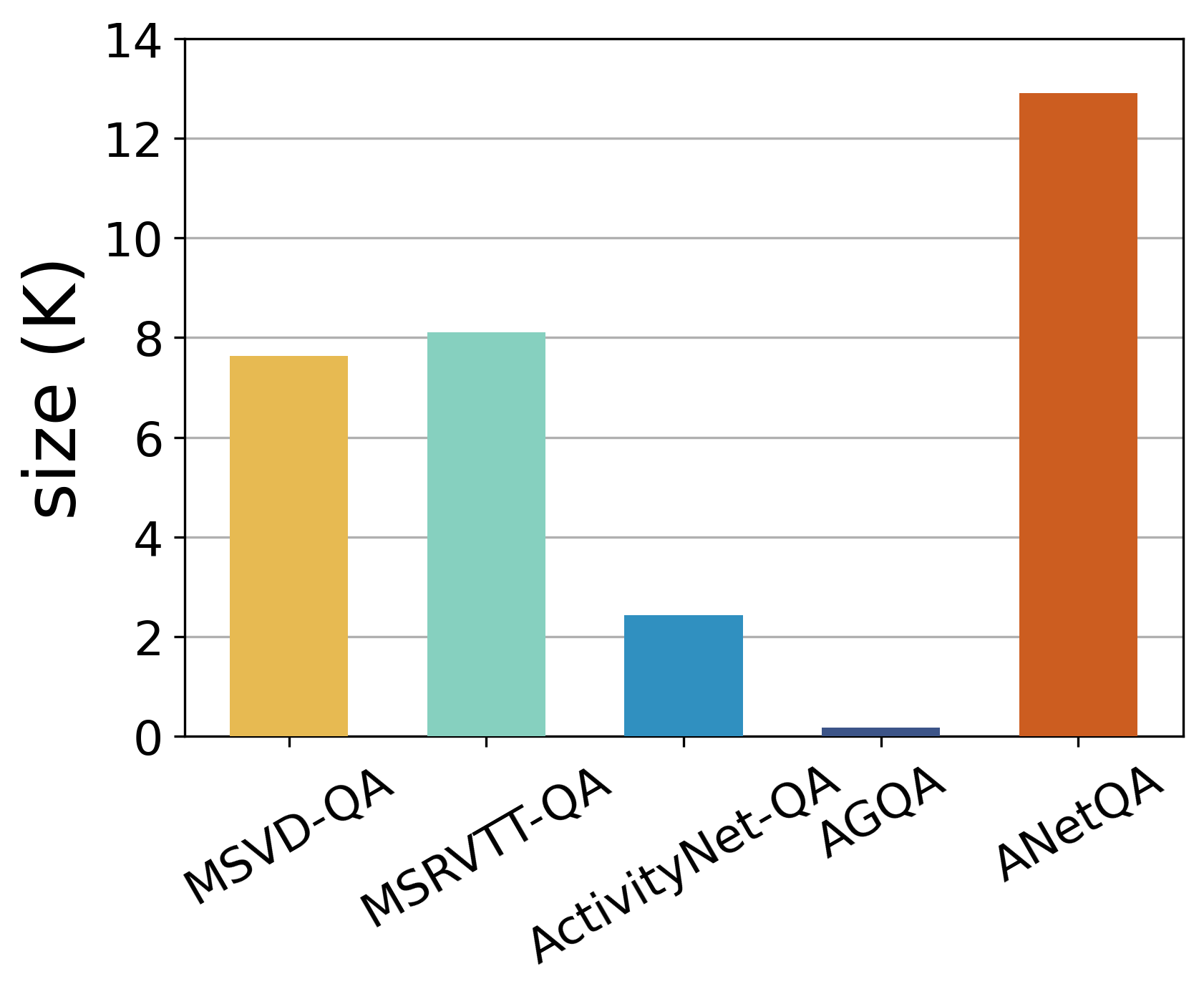}
		\caption{answer vocabulary sizes}\label{fig:alen_dist}
	\end{subfigure}
	\vspace{-5pt}
	\caption{\textbf{Question lengths and answer vocabulary sizes.} We compare the (a) question lengths and (b) answer vocabulary sizes of our ANetQA and some typical VideoQA benchmarks like MSVD-QA \cite{xu2017video}, MSRVTT-QA \cite{xu2017video}, ActivityNet-QA \cite{yu2019activitynet}, and AGQA \cite{grunde2021agqa}. Compared to the counterparts, our questions are longer and answer vocabulary size is larger, showing the fine granularity, diversity, and difficulty of our benchmark.}
	\label{fig:qa_length_comp}
	\vspace{-15pt}
\end{figure}

Our final ANetQA benchmark contains 13.4M balanced QA pairs, which consists of 10.4M \texttt{train}, 1.5M \texttt{val}, and 1.5M \texttt{test} QA pairs\footnote{For more efficient evaluation, we additionally provide a \texttt{test-dev} split by random sampling 0.3M QA pairs from the \texttt{test} split. Note that both the \texttt{test} and \texttt{test-dev} splits are conducted on the same video set and the evaluation for both splits are performed online.}. We compare the question and answer length distributions of ANetQA to existing VideoQA benchmarks. The results in Figure \ref{fig:qlen_dist} show that the ANetQA questions have a wider range of lengths and are longer on average than those of all the counterparts, showing the diversity and fine granularity of our questions, respectively. Moreover, according to these challenging questions, our answer vocabulary size is much larger than that of the counterparts (see Figure \ref{fig:alen_dist}), which further increases the difficulty of our benchmark.

\captionsetup[subfigure]{font=normalsize}

\section{Experiments}

This section contains comprehensive experiments and intensive analyses of ANetQA. We conduct evaluations on several state-of-the-art models and diagnose their capabilities to deal with different question structures, semantic classes, reasoning skills, and answer types, respectively. All the models are trained on the \texttt{train} split, validated on the \texttt{val} split, and evaluated on the \texttt{test} split. Furthermore, we also conduct a human evaluation to see the performance gap between the top-performing models and humans. Finally, we investigate the effects of different auxiliary annotations to model performance.

\subsection{Experimental Setup}

\noindent\textbf{Compared models.} We choose three state-of-the-art models for comparison, namely HCRN \cite{le2020hierarchical}, ClipBERT \cite{lei2021less}, and All-in-one \cite{wang2022all}. HCRN introduces a reusable conditional relation network (CRN) module and stacks multiple CRNs in depth to integrate the motion, question, and appearance features at different levels \cite{le2020hierarchical}. We use its default settings to extract 128 appearance features and 8 motion features, respectively. 

\captionsetup[subfigure]{font=normalsize}

Different from HCRN, ClipBERT and {All-in-one} are two Transformer-based models that incorporate vision-language pretraining (VLP) on a large-scale corpus. ClipBERT is pretrained on massive image-text pairs, which enables end-to-end learning by employing a sparse sampling mechanism. We adopt its official pretrained model weights as initial and then finetune the model on ANetQA using the (${4\times2}$) sampling strategy, which means 4 segments are sampled (with 2 sampled frames in each segment) at each training step. During model testing, we sample 16 frames uniformly for each video, as recommended in \cite{lei2021less}.
\begin{table*}
	\centering
	\begin{tabular}{ll|ccccccc|c}
		\multicolumn{2}{c|}{\multirow{2}{*}{taxnomy}} & \multirow{2}{*}{\makecell{type prior}} & \multicolumn{2}{c}{HCRN \cite{le2020hierarchical}} & \multicolumn{2}{c}{ClipBERT \cite{lei2021less}} & \multicolumn{2}{c|}{All-in-one \cite{wang2022all}} & \multirow{2}{*}{human}\\
		&& &w/ &w/o&w/ &w/o &w/ &w/o \\
		\ChangeRT{1.3pt}
		\multirow{5}{*}{\makecell{question structures}}  &query & 1.04&21.30&19.24 &23.93&16.87&\textbf{25.10}	&18.40&92.92\\
		&compare & 49.70 &	\textbf{55.66} &	50.01 & 55.62&50.06&54.41&	50.06&81.34\\
		&choose &  29.13 &	63.97 &	67.37 & 69.51&66.17&\textbf{70.39}	&67.00&71.84\\
		&verify & 50.00 	& 68.56 &	50.02 & \textbf{72.57}&50.00&72.35&50.00 &86.69\\
		&logic &50.00 &	78.70 &	76.82 & 80.06 &74.33&\textbf{80.58}	&74.20& 86.06\\
		\hline
		\multirow{4}{*}{\makecell{question semantics}}  &object & 17.74 & 55.99 &49.55 & 58.69&48.22&\textbf{59.81}&	48.99& 84.26\\
		&relationship & 22.61 &	39.65 &	33.28 & 40.19&30.89&\textbf{40.78}	&32.64&90.79\\
		&attribute & 14.60 &	35.80 &	34.05  & 39.71&32.81&\textbf{40.14}	&33.39&82.17\\
		&action &  47.83 &	72.50 &	50.29 	& \textbf{74.96}&50.99&74.39	&51.14&82.33\\
		\hline
		\multirow{6}{*}{\makecell{reasoning skills}}  &object-relationship & 10.48&	35.17&	32.38	& 37.66 & 30.03&\textbf{38.42}&	31.32&86.47\\
		&object-attribute & 17.44 &	40.95 &	37.02 & 43.72&35.45&\textbf{44.33}&	36.39&84.75\\
		&duration-comparison &50.00 &	49.90 &	49.38 & 49.98&50.10&\textbf{51.65}&54.34&76.73\\
		&exist & 50.00 &	71.20 &	56.97& \textbf{74.51}&56.31&74.49	&56.28&86.52\\
		&sequencing &  10.21 	&31.70 &	31.36 & 34.19&28.76&\textbf{35.27}&30.10&87.50\\
		&superlative & 30.32 &	47.46 &	39.78 & 49.55&38.83&\textbf{50.14}	&39.60&90.14\\	 
		\hline
		\multirow{2}{*}{answer types}  &binary &49.96 &	64.36 &	53.91 & \textbf{66.19}&53.55&65.65&53.54&83.72\\
		& open & 6.49 &	29.95	& 29.00 & 33.17&26.86&\textbf{34.33}&28.25&84.82\\
		\hline
		\hline
		\multicolumn{2}{c|}{\textbf{{overall}}}  & 17.66 &	41.15 &	37.11 & 43.92&35.55&\textbf{44.53}&36.48&	84.48	\\	  
	\end{tabular}
	\caption{\textbf{A comprehensive comparison of three VideoQA methods on ANetQA}. All results are evaluated on the \texttt{test} set. Apart from the overall accuracy, we follow \cite{grunde2021agqa} to report the per-type accuracies under different taxonomies. For each method, the variant trained with vision clues (w/) outperforms its blind counterpart without vision clues (w/o), implying that the language biases are well controlled.
	} 
	\label{table:sota_comp}
	\vspace{-5pt}
\end{table*}
{All-in-one} is a current top-performing VideoQA model, which is the first attempt to perform end-to-end video-language pretraining using raw video and textual signals as inputs \cite{wang2022all}. It is pretrained directly on a large-scale video-text corpus. We finetune its base model {All-in-one}-B on ANetQA by randomly sampling 3 frames for each video at each training step. At inference time, we also extract 3 frames uniformly and feed them to the learned model to predict the answer. 
\vspace{5pt}
\\
\noindent\textbf{Human evaluation.} We conduct an intensive human evaluation to quantify the errors and ambiguities induced during the construction of ANetQA. As the labeling costs is unaffordable to provide a thorough evaluation over all the QA pairs, we follow \cite{grunde2021agqa,hudson2019gqa} to randomly sample 4,000 QA pairs from the \texttt{test} set with the following two rules: (i) each video contains at least one sample, and (ii) each question type contains at least 50 samples. Each sample is assigned to five random annotators from a diverse group to answer the question and the majority vote over their predictions is regarded as the final human answer. 

The human performance reach at 84.48\% on the sampled \texttt{test} set. We take a closer look into these 15.52\% inconsistent human predictions and find that they are constituted by 0.75\% annotation errors, 1.95\% answer ambiguities, and 12.82\% human errors. These results imply that both of our scene graphs and generated QA pairs are of high quality. Furthermore, our benchmark contains difficult questions that even educated humans can not answer correctly. 
More analyses are provided in the supplementary material.
  
\subsection{Main Results}

We provide an intensive comparison of the state-of-the-art methods on ANetQA In Table \ref{table:sota_comp}. Besides the overall accuracy, we follow \cite{grunde2021agqa} to report the per-type accuracies under different taxonomies, \emph{i.e.}, question structures, question semantics, reasoning skills, and answer types. More detailed descriptions of the taxonomies and corresponding question templates are provided in the supplementary material. For each type, we provide a simple baseline, \emph{type prior}, that uses the most frequent answer as the prediction.

From the results, we have the general observations as follows: (i) The All-in-one model pretrained on large video-text corpus achieves the overall best performance while using the least number of sampled frames. This suggests good video representations play a central role in VideoQA performance; (ii) the best performing model is still far from the human level, showing the difficulty of our benchmark and sufficient room for further improvements; and (iii) for each method, the variant trained with vision clues (w/) steadily outperforms its \emph{blind} counterpart without any vision clues (w/o), indicating that the language biases are well controlled by our balancing strategies. 

The observations above are quite different from those on AGQA, where on their benchmark all models are on par with their corresponding blind counterparts. This can be explained that ANetQA has more unbalanced QA samples than AGQA, thus providing more room to perform thorough balancing strategies. Moreover, given the same model HCRN, its accuracy (especially the \emph{open} answer type) on ANetQA is much lower than that on AGQA, verifying the fine-grained nature of our scene graphs elements. 
\vspace{5pt}
\\
\noindent\textbf{Question structures and answer types.} The \emph{query} type questions are the most challenging ones as they have open answers. Among the rest four types which have limited answer choices\footnote{The \emph{compare}, \emph{verify}, and \emph{logic} type questions have binary answers. The \emph{choose} type question conducts a comparison between [A] and [B], and the answer refers to one of the four choices: [A], [B], both, or none.}, the \emph{compare} type questions report the lowest accuracy as they require more reasoning steps. 
\vspace{5pt}
\\
\noindent\textbf{Question semantics.} The attribute-oriented questions are the most difficult ones, as they require a more fine-grained understanding of video contents than the rest questions. 
\vspace{5pt}
\\
\noindent\textbf{Reasoning skills.} Similar to AGQA, each of our question is associated with one or more reasoning abilities necessary to answer the question. The questions requiring the \emph{sequencing} skills deliver the lowest accuracy as they require the temporal grounding ability. In contrast to the coarse action labels used in AGQA, our actions are depicted in natural language, which are more difficult to understand.

\subsection{Effects of Auxiliary Annotations}

All the comparative studies above only use the basic annotations (\emph{i.e.}, the QA pairs) for model training. As all the QA pairs are automatically generated from scene graph annotations, it is natural to investigate whether and how auxiliary annotations facilitate model performance. We introduce two auxiliary annotations \emph{scene graph statistics} and \emph{oracle frames} to see their impacts on model performance, respectively. The results are provided in \mbox{Table \ref{table:abla}}.
\vspace{5pt}
\\
\noindent\textbf{Scene graph statistics.} The annotated scene graph of a given video contains all the necessary information to answer any questions on the video. Therefore, it is meaningful to investigate the impact of this information on model performance. The fine-grained characteristics of our scene graphs make it nontrivial to encode each scene graph into a feature bank like \cite{ji2020action}. Alternatively, we introduce a simple statistical-based strategy to approximately represent the scene graph to a given video by extracting the top-$K$ high-frequency (HF) words from \emph{all} the questions on this video. The extracted HF words can be seamlessly used in any off-the-shelf model by concatenating them with the question words. We adopt HCRN \cite{le2020hierarchical} as the reference model and extract the top-40 HF words from different vocabularies (\emph{i.e.}, objects, relationships, attributes, and their combinations). These HF words are concatenated with the question words in both the training and testing phases.

From the results in the upper part of Table \ref{table:abla}, we can see that adding HF objects or relationships solely do not bring further improvement over the reference model. This can be explained by the fact that relationships are strongly coupled with objects, using either of them solely can not provide sufficient scene graph information for the model to understand. Moreover, the model with HF attributes results in a distinct performance gain compared to the counterpart with HF objects. This observation verifies that our questions requires the abilities of fine-grained understanding and reasoning. Finally, exploiting all three types of HF information results in the best performance due to their complementary nature.
\vspace{5pt}
\\
\noindent\textbf{Oracle frames.} As each question in ANetQA is generated from the scene graph elements in specific video frames, we denote these frames as the oracle frames for the question and investigate whether they can facilitate model performance. For each question, we inject the corresponding oracle frames into its sampled frames to ensure the necessary visual information to answer this question is provided. We use All-in-one \cite{wang2022all} as the reference model since it uses few sampled frames and thus has a high probability of not covering the oracle frames. We have experimented with the oracle frames in the training, testing, and both phases, respectively. The results in the lower part of \mbox{Table \ref{table:abla}} show that injecting oracle frames in the training and testing phases bring 0.87 and 0.46 point improvements over the reference model in terms of overall accuracy, respectively. Moreover, when oracle frames are applied to both the training and testing phases, the model performance is further improved due to their synergistic effects. 
\begin{table}
	\centering
	\begin{tabular}{l|ccc}
		 & binary & open & overall\\
		\ChangeRT{1.3pt}
		{(a) \textit{scene graph statistics}}\\
		HRCN \cite{le2020hierarchical} (reference) & 64.36 & 29.95 & 41.15\\
		+ high-freq. objects (O) & 65.81	&29.29&	41.18 \\
		+ high-freq. relationships (R) & 63.84	&29.21	&40.48 \\
		+ high-freq. attributes (A)& 67.67	&32.21	&43.75 \\	
		+ high-freq. O+R+A & \textbf{68.15} & \textbf{34.50} & \textbf{45.45} \\				
		\hline
		\hline
		{(b) \textit{oracle frames}}\\
		All-in-one \cite{wang2022all} (reference) & 65.65&34.33&	44.53\\
		+ training phase injection & 66.54	&35.18	&45.40\\
		+ testing phase injection & 66.04	&34.83	&44.99 \\
		+ both phases injections&  \textbf{66.88} &\textbf{36.02} & \textbf{46.07}\\
	\end{tabular}
	\caption{\textbf{Effects of different auxiliary annotations.} (a) The scene graph statistics of a given video are represented as a set of high-frequency words extracted from all the questions of that video. (b) The oracle frames contain necessary visual information to answer a given question, which are injected in different phases.}
\label{table:abla}
\vspace{-10pt}
\end{table}
\section{Conclusion and Future Work}
In this paper, we present ANetQA, a challenging VideoQA benchmark that examines fine-grained compositional reasoning over untrimmed real-world videos. Benefiting from the fine-grained video scene graphs annotated by humans, ANetQA attains 13.4M balanced QA pairs, which is an order of magnitude larger than all previous VideoQA benchmarks. We provide comprehensive experiments and intensive analyses for state-of-the-art VideoQA methods, and the best-performing model showing that a fine-grained video understanding plays a vital role in our benchmark. Moreover, there remains a significant gap between the best model and humans, indicating the challenge of our benchmark while providing room for future improvements.

We will persistently improve our benchmark. \emph{e.g.}, further reducing the language biases and answer ambiguities, and introducing more question types with diverse reasoning skills like scene-text understanding and causality inference. We hope that our ANetQA will serve as a cornerstone to facilitate future research in the video-language learning.

\section*{Acknowledgment.} 
This work was supported in part by the NSFC (61836002, 62125201), in part by the Fundamental Research Funds for the Provincial Universities of Zhejiang (GK229909299001-001), in part by the NSFC (62072147, 62020106007), and in part by the Zhejiang Provincial Natural Science Foundation of China (LR22F020001, DT23F020007).



\appendix
\section{Scene Graph Annotations}\label{app:sgstat}
\subsection{Annotation Pipeline}
As mentioned in the main paper, ANetQA is built upon the annotations of ANet-Entities \cite{zhou2019grounded}, which grounds objects in representative frames with noun phrases (NPs). Nouns and adjectives are extracted from these NPs using the Stanford Parser \cite{manning2014stanford} to form our initial object and attribute vocabularies, respectively. Meanwhile, we handcraft the initial relationship vocabulary on the activity labels of the original ActivityNet \cite{caba2015activitynet}. These initial vocabularies are intermittently updated during the annotation process. 

\begin{figure*}[h]
	\begin{center}
		\includegraphics[width=0.88\textwidth]{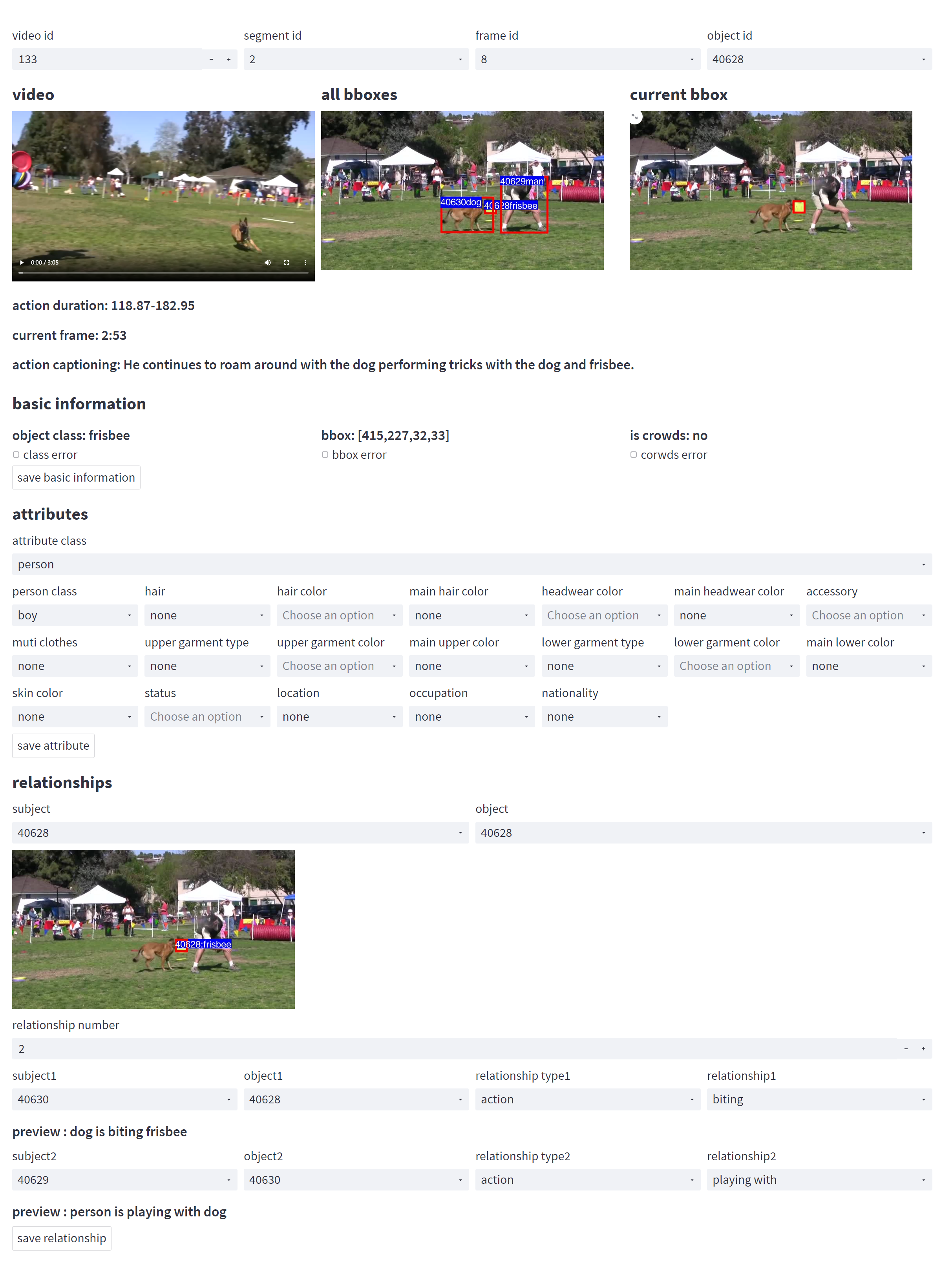}
		\caption{A web-based interface for video scene graph annotation by crowdsourcing. Annotators are asked to watch the video first and then select attributes and relationships from corresponding vocabularies. When no suitable item is available, they can add new items freely. These new items will be manually checked and the valid ones will be appended to the vocabularies intermittently. }
		\vspace{-25pt}
		\label{fig:interface}
	\end{center}
\end{figure*}

We provide a web-based interface shown in Figure \ref{fig:interface} for crowdsourcing. In total, more than 50 human annotators have participated in the annotation process for over 4 months. Each annotator is asked to watch the video first and then select attributes, and relationships from the corresponding vocabularies. When no suitable option is available, they are allowed to add a new option. These new options will be manually checked and the valid ones will be added to the vocabularies intermittently. Meanwhile, the mislabeled objects and inaccurate object bounding boxes are fixed and omitted key objects are complemented during the annotation process. To control the annotation costs, we set the maximum number of augmented objects to three.
\subsection{Scene Graph Taxonomies}
\begin{figure*}[h]
	\begin{center}
		\includegraphics[width=\textwidth]{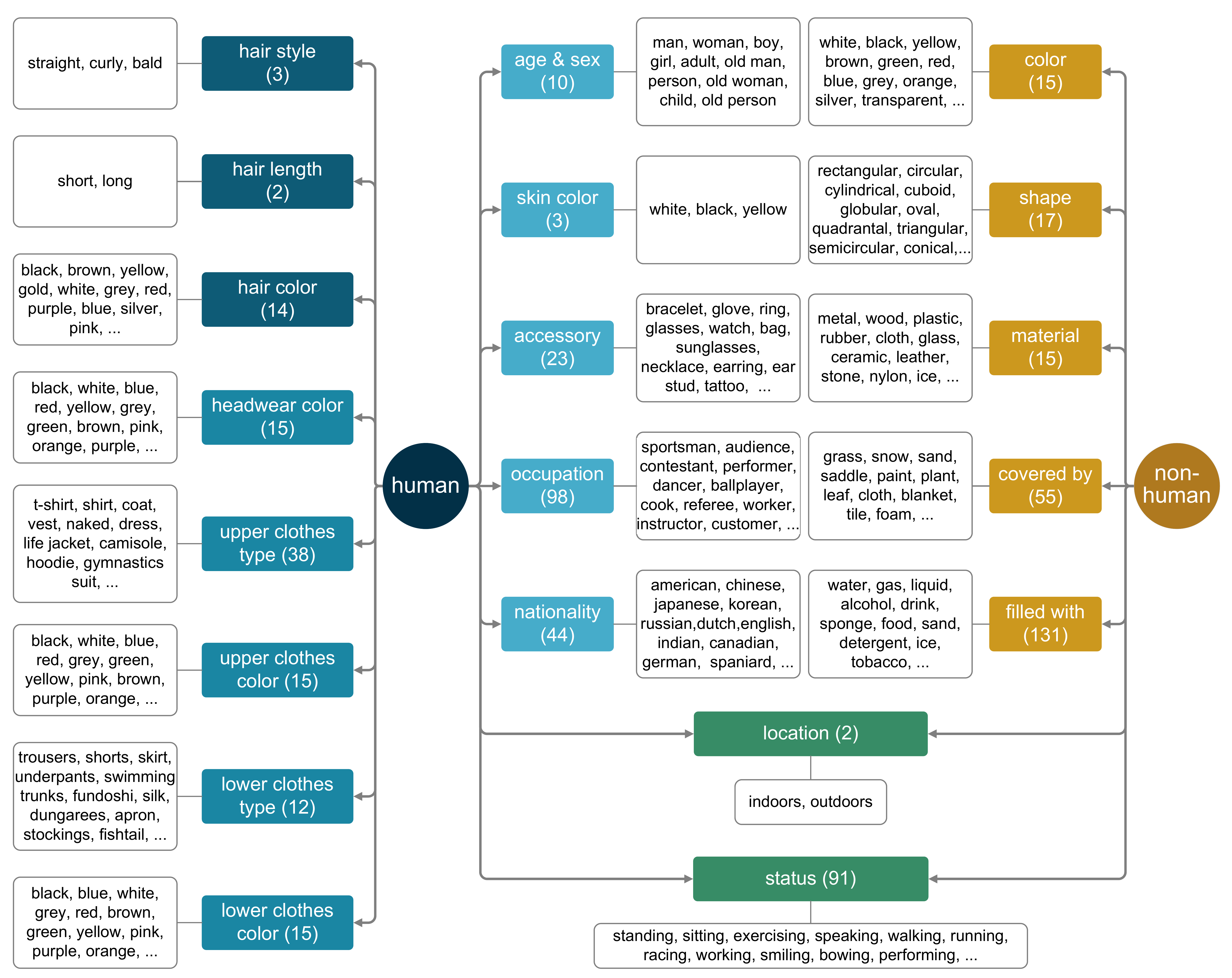}
		\vspace{-10pt}
		\caption{A hierarchy of attributes in our benchmark. The hierarchy consists of three levels. On the \textbf{top} level, objects are classified into the \emph{human} and \emph{non-human} groups. On the \textbf{middle} level, up to 20 representative attribute types are designed for each top groups (\emph{e.g.}, ``\emph{hair style}'' and ``\emph{skin color}`` for the ``\emph{human}'' group, ``shape'' and ``material'' for the ``\emph{non-human}'' group). A few attributes like ``\emph{location}'' and ``\emph{status}'' are shared across the two groups. On the \textbf{bottom} level, a total number of 618 attribute labels are provided for all the middle-level attribute types (\emph{e.g.}, ``\emph{long hair}'' and ``\emph{short hair}'' for the ``\emph{hair length}'' attribute type). For each object, annotators are asked to label the bottom-level attributes as thoroughly as possible. Due to space limitations, we show a maximum number of 10 bottom-level attributes for each mid-level attribute type.}
		\vspace{-27pt}
		\label{fig:attr_tax}
	\end{center}
\end{figure*}

\begin{figure*}[h]
	\begin{center}
		\includegraphics[width=0.98\textwidth]{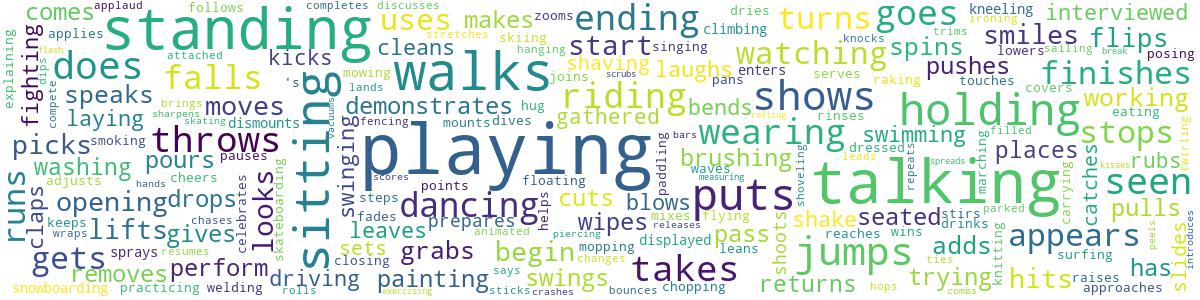}
		\caption{A word cloud for frequent \emph{verbs} in action descriptions. We merge the words with the same etymon for better visualization. 
		}
		\vspace{-27pt}
		\label{fig:act_wc}
	\end{center}
\end{figure*}

Our completed scene graph annotations include taxonomies of 2,072 object classes, 86 relationship classes, and 618 attributes classes. The detail taxonomies for objects, relationships, and attributes are shown in Table \ref{table:obj_tax}, Table \ref{table:rel_tax}, and Figure \ref{fig:attr_tax}, respectively. As our actions are depicted in natural language, we illustrate a word cloud for the most frequent verbs in Figure \ref{fig:act_wc}.

\subsection{Case Study}
\begin{figure*}[h]
	\begin{center}
		\includegraphics[width=\textwidth]{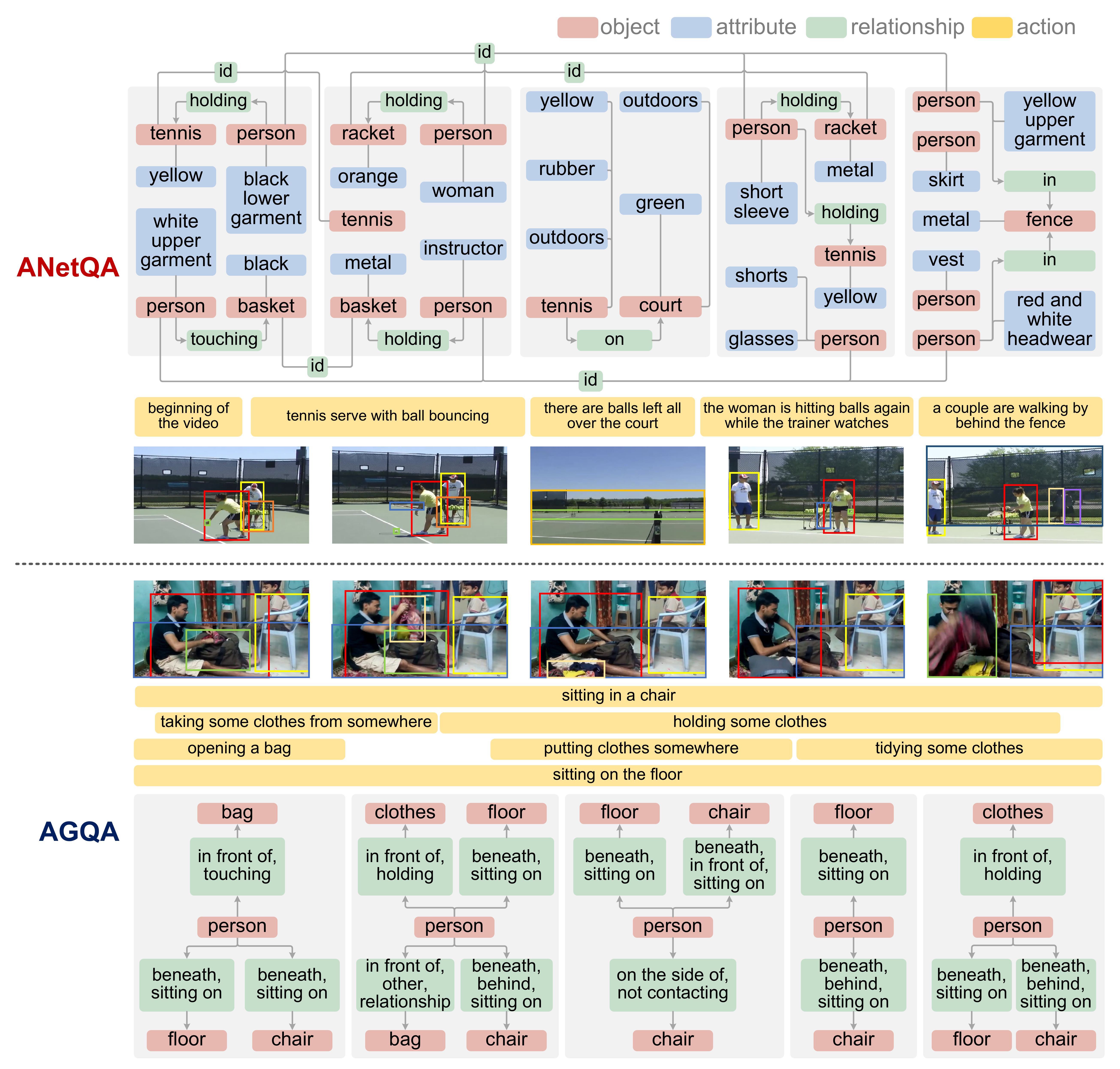}
		\caption{A comparison of the example scene graphs of our ANetQA and AGQA. The visualized results suggest: (i) our scene graph is more informative than that in AGQA as our untrimmed video contains richer semantics with multiple switched scenarios; (ii) our scene graph is much more fine-grained than that in AGQA due to the objects, relationships, actions, especially the newly introduced attributes; (iii) our scene graph contains varied relationships between human-object, human-human, and object-object pairs, while the scene graph of AGQA only contains human-object relationships; and (iv) our scene graph uses the ``\emph{identical}'' relationship to annotate the same instance across different frames, which effectively avoids the generation of ambitious questions. In contrast, the scene graph of AGQA is centered on \emph{one} person, which cannot always be satisfied in real-world videos. Specifically, the annotated ``\emph{person}'' refers to the man in the first four frames and shifts to the boy in the last frame.  
		}
		\label{fig:sg-example}
	\end{center}
\end{figure*}

In Figure \ref{fig:sg-example}, we provide comparative examples of the annotated scene graphs from ANetQA and AGQA, respectively. From the visualized results we can see that: (i) our scene graph is more informative than that in AGQA as our untrimmed video contains richer semantics with multiple switched scenarios; (ii) our scene graph is much more fine-grained than that in AGQA due to the objects, relationships, actions, especially the newly introduced attributes; (iii) our scene graph contains varied relationships between human-object, human-human, and object-object pairs, while the scene graph of AGQA only contains human-object relationships; and (iv) our scene graph uses the ``\emph{identical}'' relationship to annotate the same instance across different frames, which effectively avoids the generation of ambitious questions. In contrast, the scene graph of AGQA is centered on \emph{one} person, which cannot always be satisfied in real-world videos. As shown at the bottom, the annotated ``\emph{person}'' refers to the man in the first four frames and shifts to the boy in the last frame. 

\section{Compositional QA Generation}\label{app:qg}
\subsection{Taxonomies, Templates, and Programs}
We show the question taxonomies and templates for our benchmark in Table \ref{table:ques_template}. Similar to AGQA, each question type is categorized into different in terms of different perspectives (\emph{i.e.}, structure, semantics, reasoning skill, and answer type). Each question type corresponds to at least one question template with a maximum number of reasoning steps. Compared with AGQA, ANetQA has more diverse question templates (119 \emph{vs.} 28) , showing the diversity, fine granularity, and difficulty of our benchmark. The functional program for each template is shown in Table \ref{table:programs}.

\subsection{Question Distributions}
\begin{figure*}
	\centering
	\begin{subfigure}[h]{0.24\textwidth}
		\includegraphics[width=\linewidth]{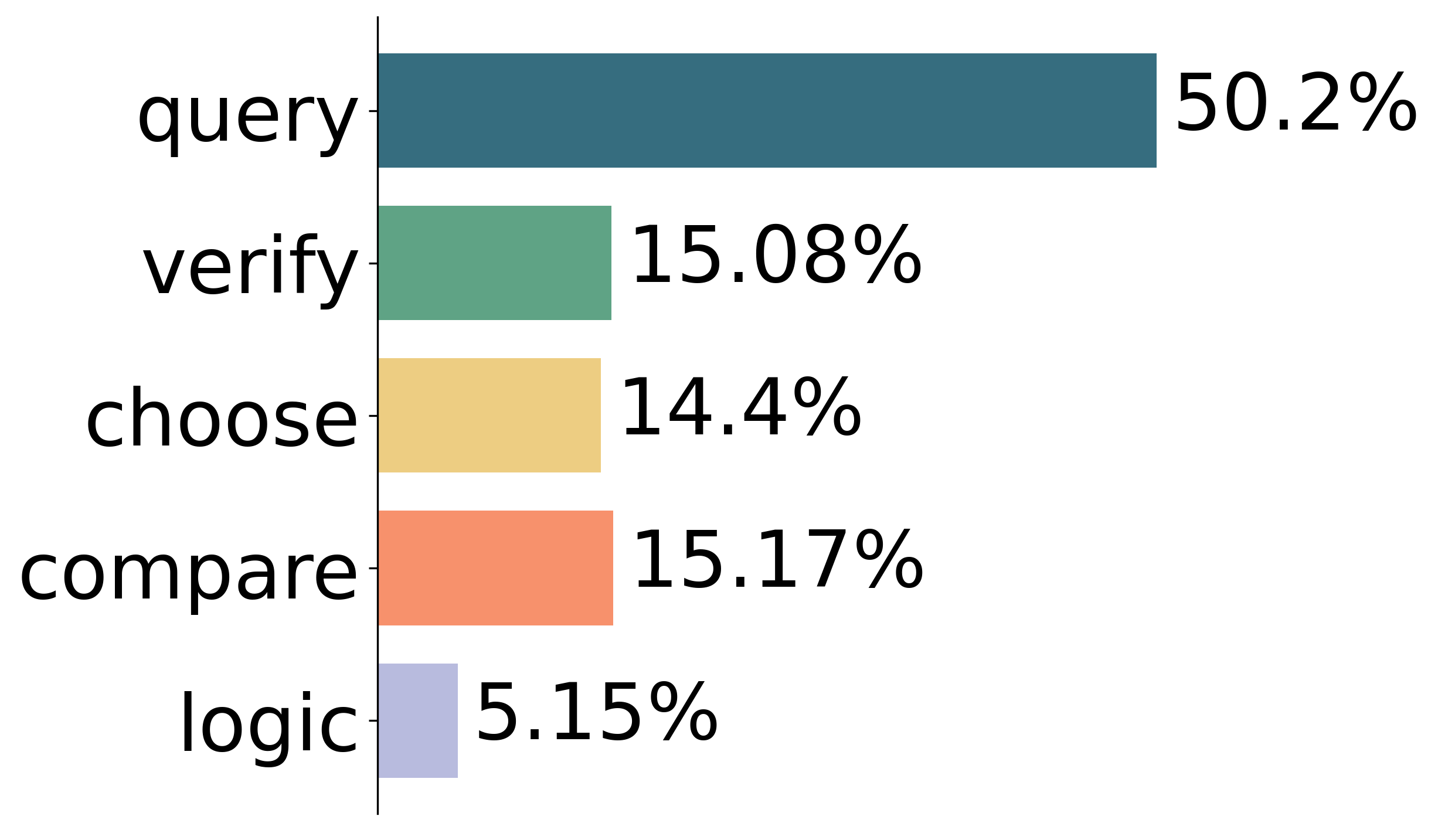}
		\vspace{-5pt}
		\caption{question structures}\label{fig:qstruct_dist}
	\end{subfigure}
	\begin{subfigure}[h]{0.24\textwidth}
		\includegraphics[width=\linewidth]{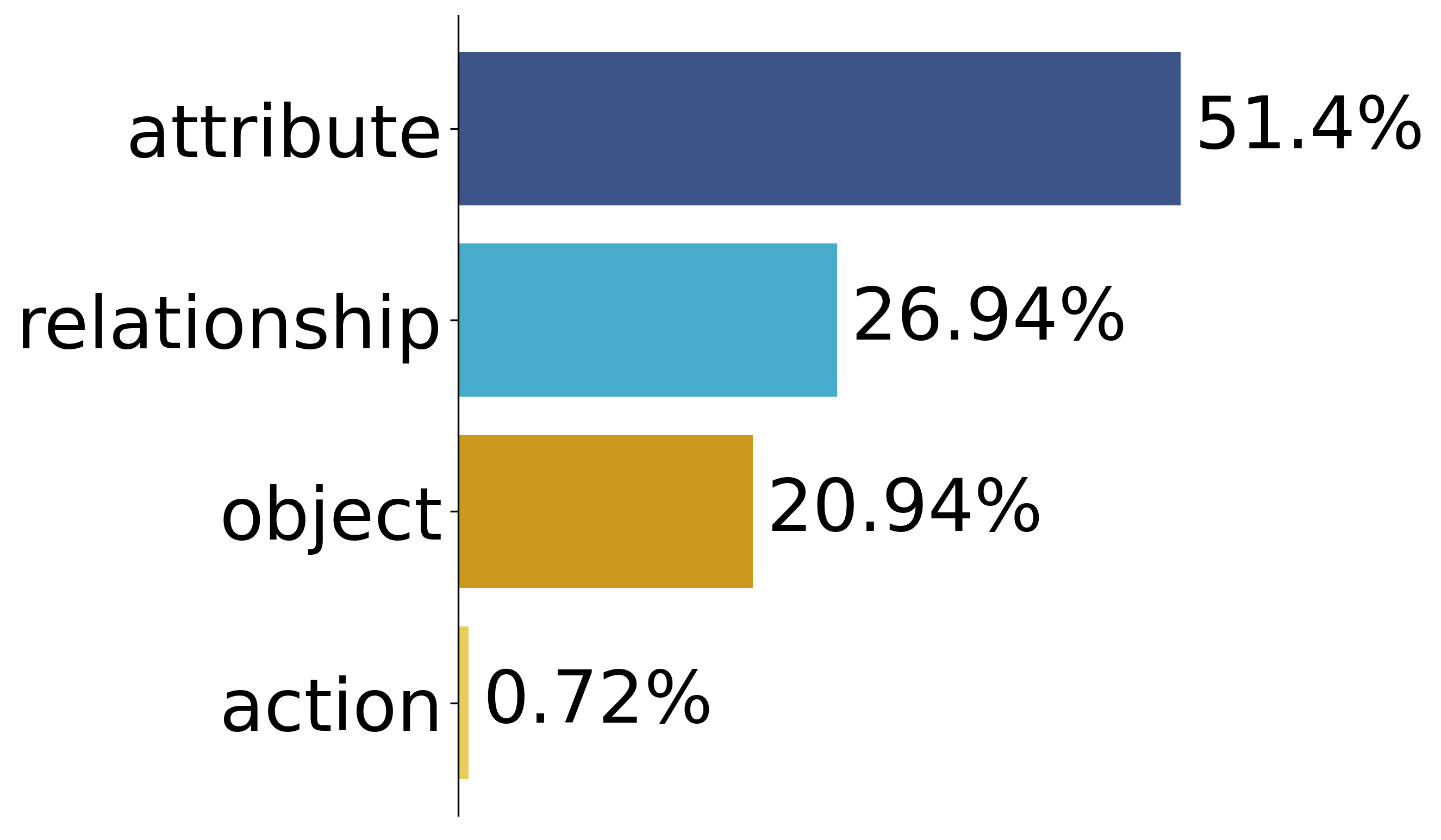}
		\vspace{-5pt}
		\caption{question semantics}\label{fig:qsem_dist}
	\end{subfigure}
	\begin{subfigure}[h]{0.27\textwidth}
		\includegraphics[width=\linewidth]{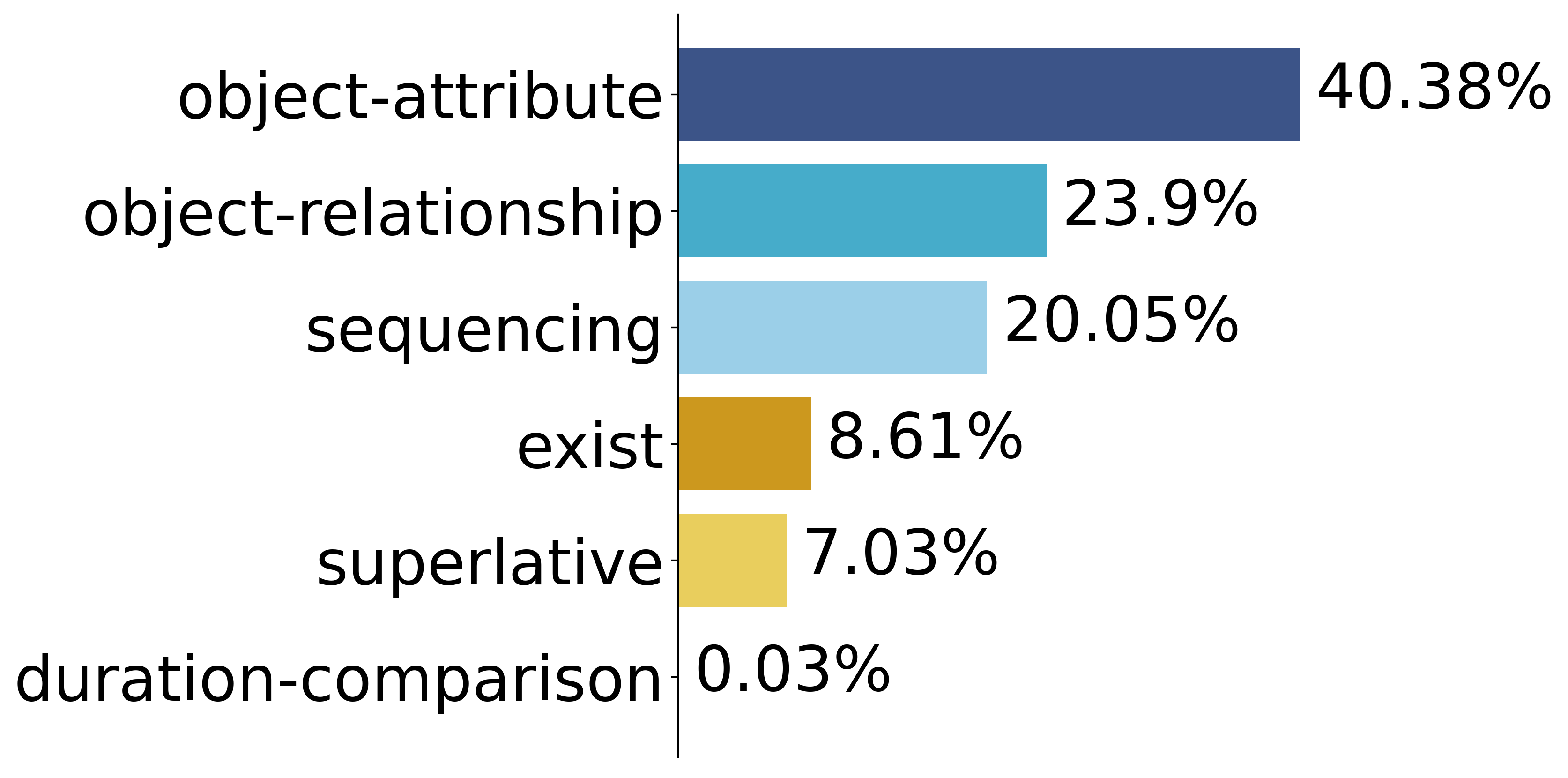}
		\vspace{-5pt}
		\caption{reasoning skills}\label{fig:reaski_dist}
	\end{subfigure}
	\begin{subfigure}[h]{0.22\textwidth}
		\includegraphics[width=\linewidth]{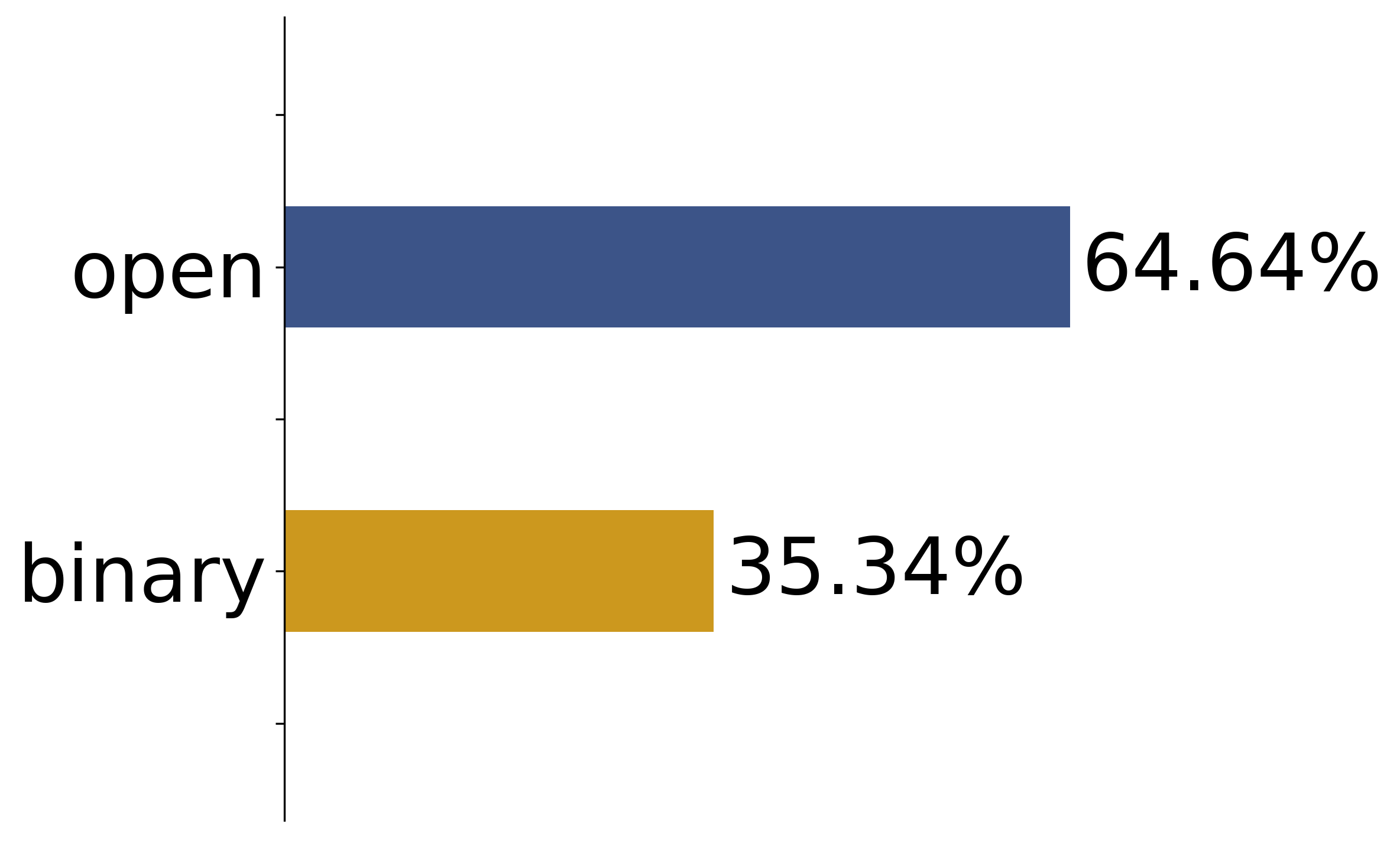}
		\vspace{-5pt}
		\caption{answer types}\label{fig:anstype_dist}
	\end{subfigure}
	\vspace{-5pt}
	\caption{\textbf{Question distributions in terms of different taxonomies} on the balanced version. (a) The question structure distribution meets the expectation of our balancing strategy; (b) and (c) The attribute-related questions account for a large percentage in terms of question semantics and reasoning skills, respectively. (d) The proportion of the \emph{open} type answers is roughly twice that of the \emph{binary} type answers.}
	\label{fig:tax_dist}
	\vspace{-5pt}
\end{figure*}

ANetQA contains 13.4M balanced QA pairs in total. We display the distributions of these QA pairs in terms of different taxonomies in Figure \ref{fig:tax_dist}. The results show that: (i) the question structure distribution meets the expectation of our balancing strategy; (ii) the attribute-related questions account for a large percentage in terms of question semantics and reasoning skills, respectively; and (iii) the proportion of the \emph{open} type answers is roughly twice that of the \emph{binary} type answers.
In Figure \ref{fig:ques_dist}, we illustrate the question distribution by the first three words. The results show that our questions are both semantically and linguistically diverse.

\subsection{Example QA pairs}
\begin{figure*}[h]
	\begin{center}
		\includegraphics[width=\textwidth]{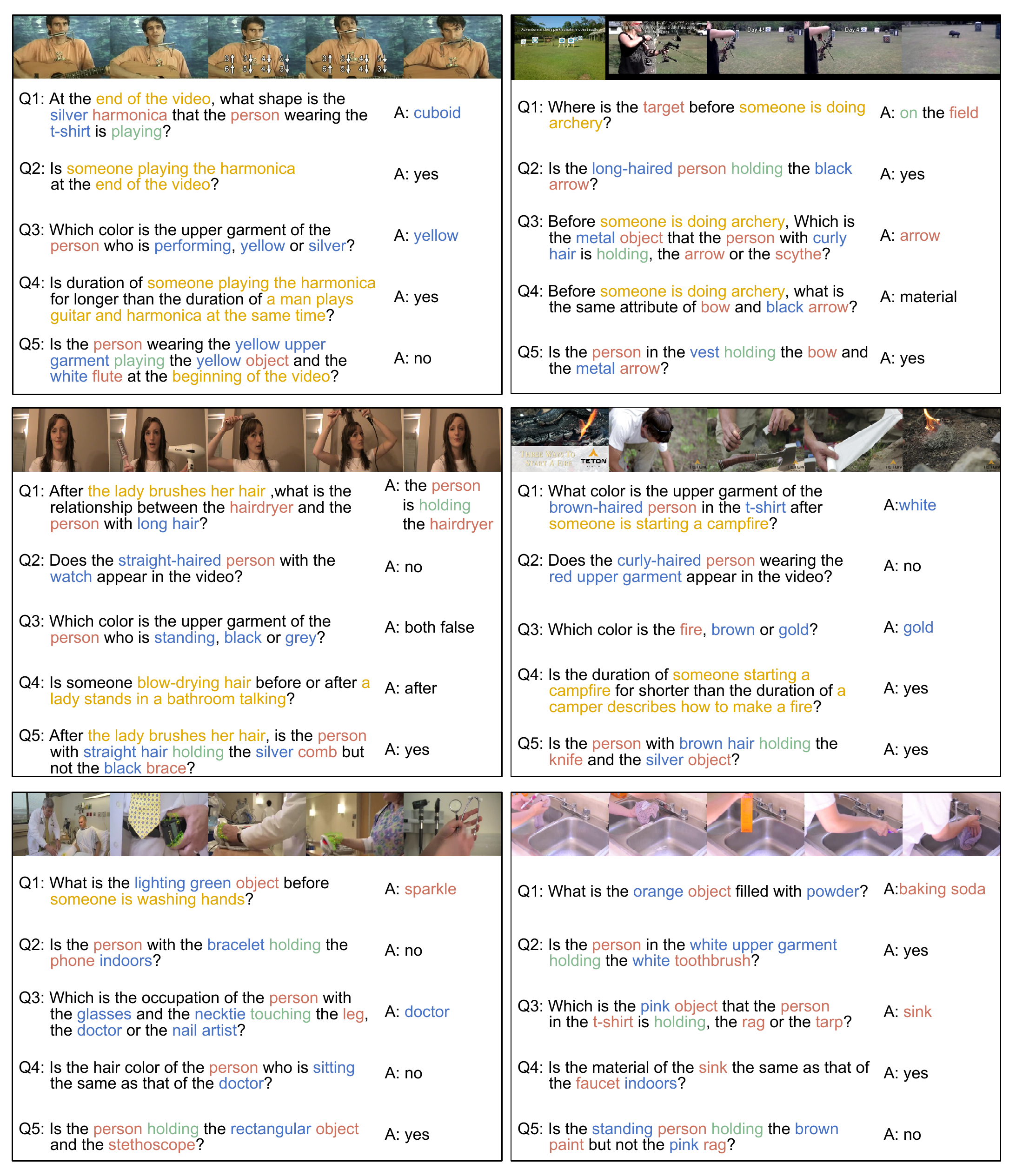}
		\caption{Example QA pairs from the \texttt{train} and \texttt{val} splits. Each example contains five QA pairs on the same video with different question structures, \emph{i.e.}, query, verify, choose, compare, and logic.}
		\label{fig:exmaple_uestion}
	\end{center}
\end{figure*}

We provide some example QA pairs from the \texttt{train} and \texttt{val} splits in Figure \ref{fig:exmaple_uestion}. Each example contains five QA pairs on the same video with different question structures (\emph{i.e.}, query, verify, choose, compare, and logic). The examples verify that our questions are diverse, fine-grained, and challenging at the same time.  

\section{Experiments}\label{app:impl}
\subsection{Human Evaluation} 
\begin{figure}
	\centering
	\begin{subfigure}[h]{0.495\columnwidth}
		\includegraphics[width=\linewidth]{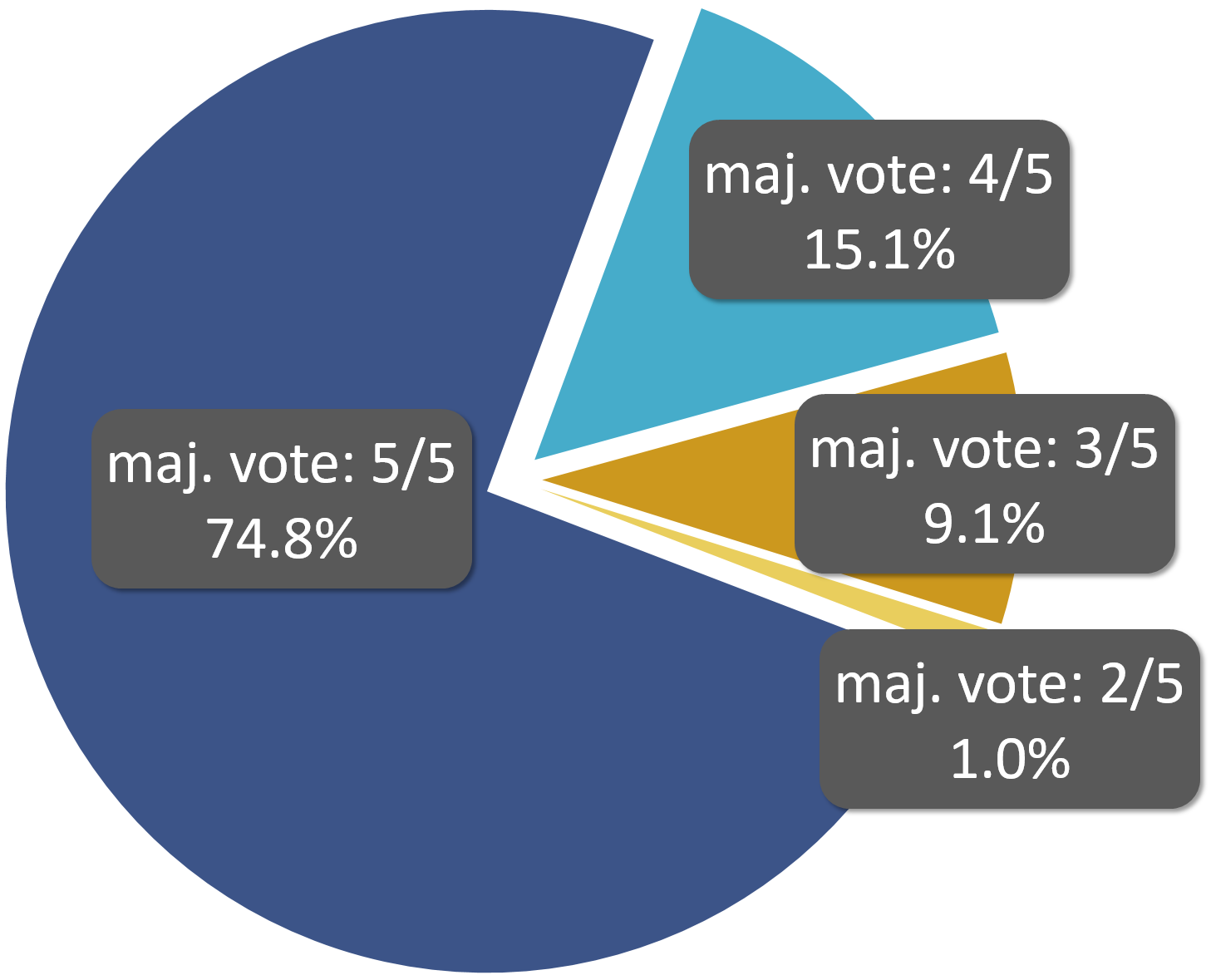}
		\vspace{-5pt}
		\caption{voting distribution}\label{fig:human-eval-vote}
	\end{subfigure}
	\begin{subfigure}[h]{0.47\columnwidth}
		\includegraphics[width=\linewidth]{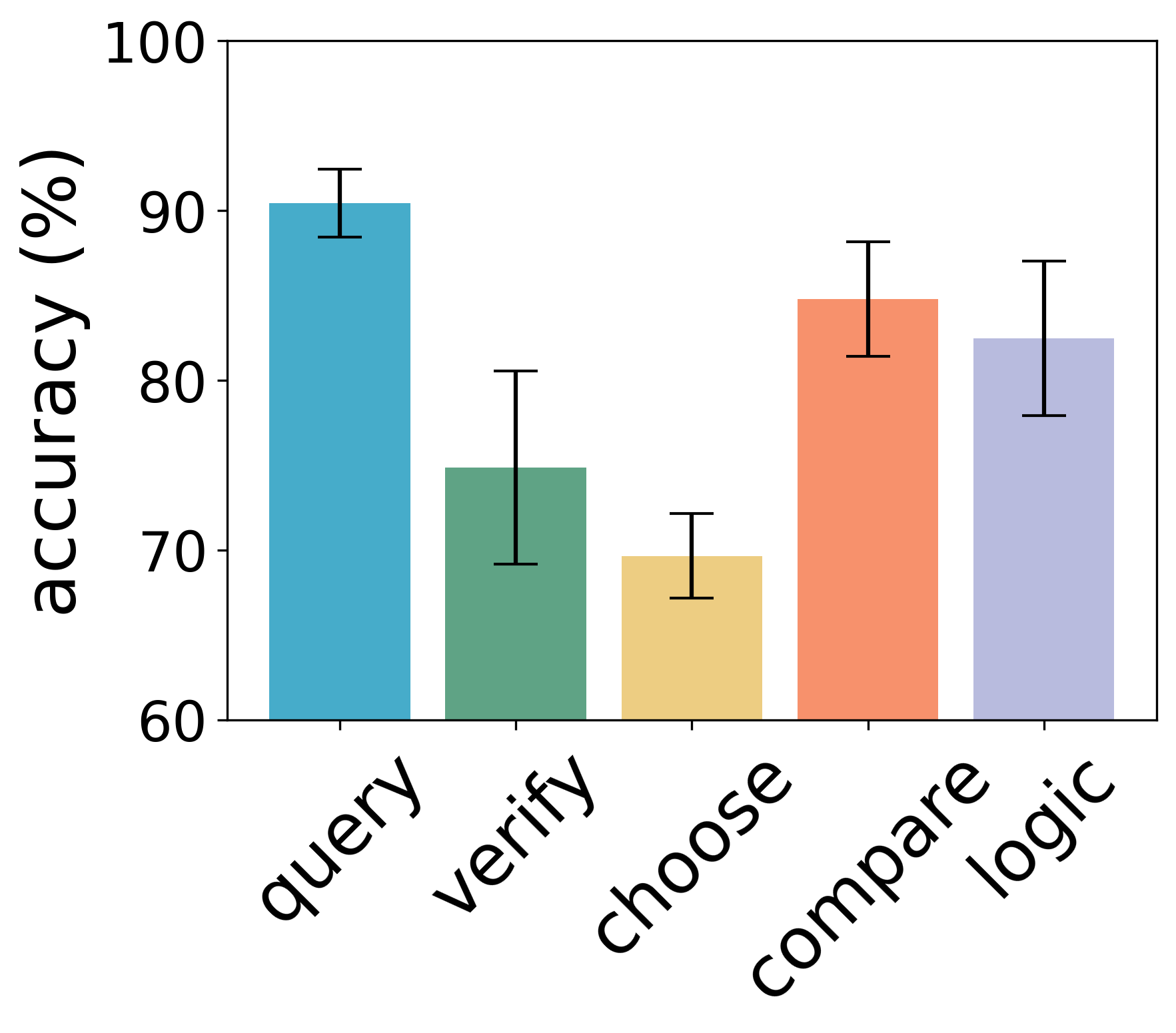}
		\vspace{-5pt}
		\caption{average accuracies}\label{fig:human-eval-structure}
	\end{subfigure}
	\vspace{-5pt}
	\caption{Given the predictions from five individual annotators, we illustrate (a) the distribution of the majority votes and (b) average accuracies with standard deviations in terms of different question structures and the overall type.}
	\label{fig:human-eval}
	\vspace{-5pt}
\end{figure}

As reported in the main paper, human performance tops out at 84.48\% overall accuracy by taking the majority voting over five answers per question. In Figure \ref{fig:human-eval}, we provide more detailed analyses of the human evaluation statistics to better understand the behavior of individual annotators. The results in Figure \ref{fig:human-eval-vote} indicate that the deviations among different annotators do exist, and majority voting helps eliminate individual errors. The results in Figure \ref{fig:human-eval-structure} show that different question types lead to diverse accuracies and deviations. 
The average accuracy per individual annotator is 81.5\%.  
\subsection{Per-Split \& Per-Type Accuracies}
In Table \ref{table:val-test}, we provide comparisons of the same model on the \texttt{val} and \texttt{test} split, respectively. The results show that: (i) the results on the \texttt{test} split is slightly lower that the \texttt{val} split; and (ii) there is no much difference between the performance on the \texttt{test} and \texttt{test-dev} splits. 

In Table \ref{table:per-type-acc}, we report the per-type accuracies of the three models. From the results we can see that the best-performing model All-in-one \cite{wang2022all} consistently outperforms the other two models in majority of the question types.  

\begin{table*}[h]
	\centering
	\begin{tabular}{|llllllll|}
		\hline
		hand&car&dog&room&water&hair&field&table\\
		horse&bike&floor&ground&river&boat&rope&board\\
		bar&wall&shoe&hill&arm&bowl&shirt&face\\
		tree&gym&pool&stage&drum&barbell&cup&skateboard\\
		track&clothes&mat&leg&snow&paper&sink&stick\\
		street&brush&tire&tool&court&beach&ingredient&head\\
		chair&glass&grass&knife&machine&roof&foot&cat\\
		wood&plate&pole&bottle&road&house&ocean&food\\
		beam&mower&bull&hoop&frisbee&yard&guitar&box\\
		window&wave&kitchen&towel&sea&pot&football&ski\\
		slope&tube&bucket&nail&bowling ball&fence&leaf&dart\\
		pumpkin&eye&canoe&pasta&building&tile&drink&rock\\
		lawn&camel&surfboard&lake&slide&rubik's cube&ice&pinata\\
		pan&contact len&kayak&counter&hat&violin&bow&pit\\
		raft&arena&fish&swing&cake&potato&cigarette&volleyball\\
		park&arrow&saxophone&baton&motorbike&croquet&racket&cookie\\
		dodgeball&carpet&bread&sandwich&short sleeves&vacuum&hockey&hammer\\
		bag&shovel&area&elliptical machine&javelin&curling&kite&shot\\
		mirror&tennis&piano&lemon&mouth&door&sidewalk&accordion\\
		line&icecream&shop&shuffleboard&table tennis&lane&stair&body\\
		microphone&finger&paint&net&harmonica&helmet&liquid&water polo\\
		discus&product&egg&bathroom&platform&fire&gun&studio\\
		suit&alcohol&back&paddle&sand&glove&mop&hole\\
		sofa&stilt&stand&pin&beer&flute&dish&rag\\
		smoke&scissors&tattoo&sky&tomato&razor&vest&basketball\\
		\hline
	\end{tabular}
	\caption{A list of top-200 object classes in terms of occurrences in our benchmark. Sorted by row first.}
	\label{table:obj_tax}
\end{table*}

\begin{table*}[h]
	\centering
	\begin{tabular}{|c|llllll|}
		\hline
		spatial& near & in & on & part of&&\\
		\hline
		~~~~~temporal~~~~~ & identical&&&&& \\
		\hline
		\multirow{14}{*}{contact}&pulling&holding&touching&fighting with&wearing&hitting\\
		&playing&standing on&playing with&sweeping&wiping&sitting on\\
		&spitting&stirring&eating&jumping into&taking picture of&driving\\
		&riding&leading&throwing&climbing&leaning on&covering\\
		&lying on&kneeling on&walking on&raising&biting&hugging\\
		&cutting&running on&jumping on&squating on&trimming&scraping\\
		&carrying&pushing&brushing&pointing at&dancing with&chasing\\
		&surfing on&polishing&washing&drinking from&stamping&fishing\\
		&speaking with&pouring&drinking&crossing&dragging&repairing\\
		&smoking&sliding on&bowing to&drawing on&hanging on&drawn on\\
		&making&flying from&drawing&feeding&poured into&flowing from\\
		&kissing&twisting&writing on&burning&lighting&pouring into\\
		&spraying&commanding&blowing&heating&pointing&painting on\\
		&painting&painted on&wirting on&&&\\
		\hline
	\end{tabular}
	\caption{{A list of all the 86 relationships} in our benchmark, including 4 spatial, 1 temporal, and 81 contact relationships. Sorted by row first in terms of occurrences.}
	\label{table:rel_tax}
\end{table*}

\begin{sidewaystable*}[h]
	\centering
	\footnotesize
	\begin{tabular}{l|cccccc|l}
		type & \makecell{question\\structures} & \makecell{question\\semantics} & \makecell{reasoning\\skill}  & \makecell{answer\\types} & \makecell{reasoning\\steps} & \#templ. & question template\\
		\ChangeRT{1.3pt}
		\multirow{2}{*}{attrRelWhat} & \multirow{2}{*}{query} & \multirow{2}{*}{attribute} & \multirow{2}{*}{obj-attr,obj-rel} & \multirow{2}{*}{open}& \multirow{2}{*}{5} & 15 & what \textcolor{attr_c2}{[attr-type]} is the \textcolor{attr_c}{[attr1]} \textcolor{obj_c}{[obj1]} \textcolor{rel_c}{\textcolor{rel_c}{[rel]}} \textcolor{attr_c}{[attr2]} \textcolor{obj_c}{[obj2]}?\\
		&&&&&& 15& what \textcolor{attr_c2}{[attr-type]} is the \textcolor{attr_c}{[attr1]} \textcolor{obj_c}{[obj1]} that \textcolor{attr_c}{[attr2]} \textcolor{obj_c}{[obj2]} is \textcolor{rel_c}{\textcolor{rel_c}{[rel]}}?\\
		\hline
		attrWhat & query &  attribute & obj-attr &open& 3 & 15 & what \textcolor{attr_c2}{[attr-type]} is the \textcolor{attr_c}{[attr]} \textcolor{obj_c}{[obj]}?\\
		\hline
		relWhat & query & relationship & obj-attr,obj-rel& open & 5&1&what is the relationship between \textcolor{attr_c}{[attr1]} \textcolor{obj_c}{[obj1]} and \textcolor{attr_c}{[attr2]} \textcolor{obj_c}{[obj2]}?\\
		\hline
		\multirow{2}{*}{objRelWhere} & 	\multirow{2}{*}{query} & 	\multirow{2}{*}{relationship} &	\multirow{2}{*}{obj-attr,obj-rel}& \multirow{2}{*}{open}& 	\multirow{2}{*}{5} & 1 & where is the \textcolor{attr_c}{[attr1]} \textcolor{obj_c}{[obj1]} \textcolor{rel_c}{[rel]} \textcolor{attr_c}{[attr2]} \textcolor{obj_c}{[obj2]}?\\
		&&&&&& 1& where is the \textcolor{attr_c}{[attr1]} \textcolor{obj_c}{[obj1]} that \textcolor{attr_c}{[attr2]} \textcolor{obj_c}{[obj2]} is \textcolor{rel_c}{[rel]}?\\
		\hline
		\multirow{2}{*}{objRelWhat} & 	\multirow{2}{*}{query} & 	\multirow{2}{*}{object} &	\multirow{2}{*}{obj-attr, obj-rel}& \multirow{2}{*}{open}& 	\multirow{2}{*}{5} & 1 & what is the \textcolor{attr_c}{[attr1]} object \textcolor{rel_c}{[rel]} \textcolor{attr_c}{[attr2]} \textcolor{obj_c}{[obj2]}?\\
		&&&&&& 1& what is the \textcolor{attr_c}{[attr1]} object that \textcolor{attr_c}{[attr2]} \textcolor{obj_c}{[obj2]} is \textcolor{rel_c}{[rel]}?\\
		\hline
		objWhere & query  & relationship & obj-attr,obj-rel &open & 3&1&where is the \textcolor{attr_c}{[attr]} \textcolor{obj_c}{[obj]}?\\
		\hline
		objWhat & query & object & obj-attr & open & 3 & 1 & what is \textcolor{attr_c}{[attr]} object?\\
		\hline
		objExist & verify & object&exists,obj-attr&binary & 3&1 & does \textcolor{attr_c}{[attr]} \textcolor{obj_c}{[obj]} appear?\\
		\hline
		objRelExist & verify & relationship & exists,obj-attr,obj-rel & binary & 5 & 1 & is \textcolor{attr_c}{[attr1]} \textcolor{obj_c}{[obj1]} \textcolor{rel_c}{[rel]} \textcolor{attr_c}{[attr2]} \textcolor{obj_c}{[obj2]}?
		\\
		\hline
		actExist & verify &action & exist & binary & 2 & 1 & is someone \textcolor{act_c}{[act]}?\\
		\hline
		\multirow{2}{*}{objRelWhatChoose} & \multirow{2}{*}{choose} &\multirow{2}{*}{object} &\multirow{2}{*}{obj-attr,obj-rel} &\multirow{2}{*}{open}& \multirow{2}{*}{5}&1&which is \textcolor{attr_c}{[attr1]} object \textcolor{rel_c}{[rel]} \textcolor{attr_c}{[attr2]} \textcolor{obj_c}{[obj2]}, \textcolor{obj_c}{[obj-A]} or \textcolor{obj_c}{[obj-B]}?\\
		&&&&&& 1& which is \textcolor{attr_c}{[attr1]} object that \textcolor{attr_c}{[attr2]} \textcolor{obj_c}{[obj2]} is \textcolor{rel_c}{[rel]}, \textcolor{obj_c}{[obj-A]} or \textcolor{obj_c}{[obj-B]}?\\
		\hline
		objWhatChoose & choose & object& obj-attr& open& 3& 1 & which is \textcolor{attr_c}{[attr]} object, \textcolor{obj_c}{[obj-A]} or \textcolor{obj_c}{[obj-B]}?\\
		\hline
		\multirow{2}{*}{attrRelWhatChoose} & \multirow{2}{*}{choose} &\multirow{2}{*}{attribute}& \multirow{2}{*}{obj-attr,obj-rel} &\multirow{2}{*}{open}& \multirow{2}{*}{5} & 18& which \textcolor{attr_c2}{[attr-type]} is the \textcolor{attr_c}{[attr1]} \textcolor{obj_c}{[obj1]} \textcolor{rel_c}{[rel]} \textcolor{attr_c}{[attr2]} \textcolor{obj_c}{[obj2]}, \textcolor{attr_c}{[attr-A]} or \textcolor{attr_c}{[attr-B]}?
		\\
		&&&&&& 18& which \textcolor{attr_c2}{[attr-type]} is the \textcolor{attr_c}{[attr1]} \textcolor{obj_c}{[obj1]} that \textcolor{attr_c}{[attr2]} \textcolor{obj_c}{[obj2]} is \textcolor{rel_c}{[rel]}, \textcolor{attr_c}{[attr-A]} or \textcolor{attr_c}{[attr-B]}?
		\\
		\hline
		attrWhatChoose & choose & attribute & obj-attr& open & 3&18& which \textcolor{attr_c2}{[attr-type]} is the \textcolor{attr_c}{[attr]} \textcolor{obj_c}{[obj]}, \textcolor{attr_c}{[attr-A]} or \textcolor{attr_c}{[attr-B]}?
		\\
		\hline
		attrCompare & compare & attribute & obj-attr& binary & 5 & 1 & is the \textcolor{attr_c2}{[attr-type]} of the \textcolor{attr_c}{[attr]} \textcolor{obj_c}{[obj]} the same as that of the \textcolor{attr_c}{[attr]} \textcolor{obj_c}{[obj]}?\\
		\hline
		attrSame & compare & attribute& obj-attr & open &5 &1 & what is the same attributes of \textcolor{attr_c}{[attr1]} \textcolor{obj_c}{[obj1]} and \textcolor{attr_c}{[attr2]} \textcolor{obj_c}{[obj2]}?
		\\
		\hline
		actTime & compare & action & suquencing & binary & 5 & 1 & is someone \textcolor{act_c}{[act]} before or after \textcolor{act_c}{[act]}?
		\\
		\hline
		actLongerVerify & compare & action & duration-comparison & binary& 5 & 1 &  is the duration of someone \textcolor{act_c}{[act1]} for longer than the duration of \textcolor{act_c}{[act2]}?\\
		\hline
		actShorterVerify & compare & action & duration-comparison & binary& 5 & 1 &  is the duration of someone \textcolor{act_c}{[act1]} for shorter than the duration of \textcolor{act_c}{[act2]}?\\
		\hline
		andObjRelExist & logic & relationship & exists,obj-attr,obj-rel & binary & 8 & 1 & is \textcolor{attr_c}{[attr1]} \textcolor{obj_c}{[obj1]} \textcolor{rel_c}{[rel]} \textcolor{attr_c}{[attr2]} \textcolor{obj_c}{[obj2]} and \textcolor{attr_c}{[attr3]} \textcolor{obj_c}{[obj3]}?\\
		\hline
		xorObjRelExist & logic & relationship & exists,obj-attr,obj-rel & binary & 8 & 1  & is \textcolor{attr_c}{[attr1]} \textcolor{obj_c}{[obj1]} \textcolor{rel_c}{[rel]} \textcolor{attr_c}{[attr2]} \textcolor{obj_c}{[obj2]} but not \textcolor{attr_c}{[attr3]} \textcolor{obj_c}{[obj3]}?\\
	\end{tabular}
	\vspace{7pt}
	\caption{{Question taxonomy and templates.} ANetQA contains 21 types of questions generated from 119 templates. Each question type is categorized into different taxonomies (\emph{i.e.}, structure, semantics, reasoning skill, and answer type), and refers to a maximum number of reasoning steps. Note that the reasoning skills of \emph{sequencing} and {superlative} are optionally used in all the question types by inserting a clause starting with ``before/after \textcolor{act_c}{[act]}'' or ``in the beginning/end of the video''. \textcolor{attr_c2}{[attr-type]} refers to a set of templates that ask different middle-level attribute types shown in Figure \ref{fig:attr_tax}. Note that some attribute types may slightly deviate  from the corresponding template (\emph{e.g.}, ``\emph{what is the occupation of ...}'' or ``\emph{what are the accessories of ...}''). Due to space limitations, we do not expand all the templates and only show the most commonly-used one for those question types with multiple templates. }\label{table:ques_template}
\end{sidewaystable*}

\begin{table*}
	\centering
	\begin{tabular}{l|l}
		template & functional program \\
		\ChangeRT{1.3pt}
		\makecell[l]{what \textcolor{attr_c2}{[attr-type]} is the \textcolor{attr_c}{[attr1]} \textcolor{obj_c}{[obj1]} \textcolor{rel_c}{\textcolor{rel_c}{[rel]}}
			\textcolor{attr_c}{[attr2]} \textcolor{obj_c}{[obj2]}?}
		&
		\multirow{3}{*}{\makecell[l]{\texttt{select}:\textcolor{obj_c}{[obj2]}$\rightarrow$\texttt{filter}:\textcolor{attr_c}{[attr2]}$\rightarrow$\texttt{relate}:\textcolor{obj_c}{[obj1]},\textcolor{rel_c}{[rel]}
				\\$\rightarrow$\texttt{filter}:\textcolor{attr_c}{[attr1]}$\rightarrow$\texttt{query}:$\langle$\textcolor{attr_c2}{[attr-type]}$\rangle$}}
		\\
		\cline{1-1}
		what \textcolor{attr_c2}{[attr-type]} is the \textcolor{attr_c}{[attr2]} \textcolor{obj_c}{[obj2]} that \textcolor{attr_c}{[attr1]} \textcolor{obj_c}{[obj1]}\\ is \textcolor{rel_c}{\textcolor{rel_c}{[rel]}}?
		&   \\
		\hline
		what \textcolor{attr_c2}{[attr-type]} is the \textcolor{attr_c}{[attr]} \textcolor{obj_c}{[obj]}? 
		& 
		\makecell[l]{\texttt{select}:\textcolor{obj_c}{[obj]}$\rightarrow$\texttt{filter}:\textcolor{attr_c}{[attr]}$\rightarrow$\texttt{query}:$\langle$\textcolor{attr_c2}{[attr-type]}$\rangle$}
		\\
		\hline
		\makecell[l]{what is the relationship between \textcolor{attr_c}{[attr1]} \textcolor{obj_c}{[obj1]} \\and \textcolor{attr_c}{[attr2]} \textcolor{obj_c}{[obj2]}?} 
		&
		\makecell[l]{\texttt{select}:\textcolor{obj_c}{[obj1]}$\rightarrow$\texttt{filter}:\textcolor{attr_c}{[attr1]}$\rightarrow$\texttt{select}: \textcolor{obj_c}{[obj2]}\\ $\rightarrow$\texttt{filter}:\textcolor{attr_c}{[attr2]}$\rightarrow$\texttt{query}:$\langle${relationship}$\rangle$} \\
		\hline
		where is the \textcolor{attr_c}{[attr1]} \textcolor{obj_c}{[obj1]} \textcolor{rel_c}{[rel]} \textcolor{attr_c}{[attr2]} \textcolor{obj_c}{[obj2]}?
		&
		\multirow{2}{*}{\makecell[l]{\texttt{select}:\textcolor{obj_c}{[obj2]}$\rightarrow$\texttt{filter}:\textcolor{attr_c}{[attr2]}$\rightarrow$\texttt{relate}:\textcolor{obj_c}{[obj1]},\textcolor{rel_c}{[rel]}\\
				$\rightarrow$\texttt{filter}:\textcolor{attr_c}{[attr1]}$\rightarrow$\texttt{query}:$\langle$spatial-relationship$\rangle$}}   
		\\
		\cline{1-1}
		where is the \textcolor{attr_c}{[attr1]} \textcolor{obj_c}{[obj1]} that \textcolor{attr_c}{[attr2]} \textcolor{obj_c}{[obj2]} is \textcolor{rel_c}{[rel]}? & \\
		\hline
		what is the \textcolor{attr_c}{[attr1]} object \textcolor{rel_c}{[rel]} \textcolor{attr_c}{[attr2]} \textcolor{obj_c}{[obj2]}?
		&    
		\multirow{2}{*}{\makecell[l]{\texttt{select}:\textcolor{obj_c}{[obj2]}$\rightarrow$\texttt{filter}:\textcolor{attr_c}{[attr2]}$\rightarrow$\texttt{relate}:\_,\textcolor{rel_c}{[rel]}\\
				$\rightarrow$\texttt{filter}:\textcolor{attr_c}{[attr1]}$\rightarrow$\texttt{query}:$\langle$object$\rangle$}}   
		\\
		\cline{1-1}
		what is the \textcolor{attr_c}{[attr1]} object that \textcolor{attr_c}{[attr2]} \textcolor{obj_c}{[obj2]} is \textcolor{rel_c}{[rel]}? & 
		\\
		\hline
		where is the \textcolor{attr_c}{[attr]} \textcolor{obj_c}{[obj]}? 
		&
		\makecell[l]{\texttt{select}:\textcolor{obj_c}{[obj]}$\rightarrow$\texttt{filter}:\textcolor{attr_c}{[attr]}$\rightarrow$\texttt{query}:$\langle$spatial-relationship$\rangle$}	 
		\\
		\hline
		what is \textcolor{attr_c}{[attr]} object?
		&
		\makecell[l]{\texttt{select}:\_$\rightarrow$\texttt{filter}:\textcolor{attr_c}{[attr]}$\rightarrow$\texttt{query}:$\langle$object$\rangle$}
		\\
		\hline
		does \textcolor{attr_c}{[attr]} \textcolor{obj_c}{[obj]} appear?
		&
		\makecell[l]{\texttt{select}:\textcolor{obj_c}{[obj]}$\rightarrow$\texttt{filter}:\textcolor{attr_c}{[attr]}$\rightarrow$\texttt{exist}}
		\\
		\hline
		is \textcolor{attr_c}{[attr1]} \textcolor{obj_c}{[obj1]} \textcolor{rel_c}{[rel]} \textcolor{attr_c}{[attr2]} \textcolor{obj_c}{[obj2]}?
		&
		\makecell[l]{\texttt{select}:\textcolor{obj_c}{[obj1]}$\rightarrow$\texttt{filter}:\textcolor{attr_c}{[attr1]}$\rightarrow$\texttt{relate}:\textcolor{obj_c}{[obj2]},\textcolor{rel_c}{[rel]} \\
			$\rightarrow$\texttt{filter}:\textcolor{attr_c}{[attr2]}$\rightarrow$\texttt{exist}}
		\\
		\hline
		is someone \textcolor{act_c}{[act]}?
		&
		\makecell[l]{\texttt{select}:\textcolor{act_c}{[act]}$\rightarrow$\texttt{exist}}
		\\
		\hline
		\makecell[l]{which is \textcolor{attr_c}{[attr1]} object \textcolor{rel_c}{[rel]} \textcolor{attr_c}{[attr2]} \textcolor{obj_c}{[obj2]}, \\ \textcolor{obj_c}{[obj-A]} or \textcolor{obj_c}{[obj-B]}?}
		&
		\multirow{3}{*}{\makecell[l]{\texttt{select}:\textcolor{obj_c}{[obj2]}$\rightarrow$\texttt{filter}:\textcolor{attr_c}{[attr2]}$\rightarrow$\texttt{relate}:\_,\textcolor{rel_c}{[rel]}\\$\rightarrow$\texttt{filter}:\textcolor{attr_c}{[attr1]}$\rightarrow$\texttt{choose}:\textcolor{obj_c}{[obj-A]} $\mid$ \textcolor{obj_c}{[obj-B]}}}
		\\
		\cline{1-1}
		\makecell[l]{which is \textcolor{attr_c}{[attr1]} object that \textcolor{attr_c}{[attr2]} \textcolor{obj_c}{[obj2]} is \textcolor{rel_c}{[rel]}, \\ \textcolor{obj_c}{[obj-A]} or \textcolor{obj_c}{[obj-B]}?} &
		\\
		\hline
		which is \textcolor{attr_c}{[attr]} object, \textcolor{obj_c}{[obj-A]} or \textcolor{obj_c}{[obj-B]}?
		&
		\makecell[l]{\texttt{select}:\_$\rightarrow$\texttt{filter}:\textcolor{attr_c}{[attr]}$\rightarrow$\texttt{choose}:\textcolor{obj_c}{[obj-A]} $\mid$ \textcolor{obj_c}{[obj-B]}} 
		\\
		\hline
		\makecell[l]{which \textcolor{attr_c2}{[attr-type]} is the \textcolor{attr_c}{[attr1]} \textcolor{obj_c}{[obj1]} \textcolor{rel_c}{[rel]} \textcolor{attr_c}{[attr2]} \textcolor{obj_c}{[obj2]}, \\\textcolor{attr_c}{[attr-A]} or \textcolor{attr_c}{[attr-B]}?}
		&  \multirow{3}{*}{\makecell[l]{\texttt{select}:\textcolor{obj_c}{[obj2]}$\rightarrow$ \texttt{filter}:\textcolor{attr_c}{[attr2]}$\rightarrow$\texttt{relate}:\textcolor{obj_c}{[obj1]},\textcolor{rel_c}{[rel]}\\$\rightarrow$\texttt{filter}\textcolor{attr_c}{[attr1]}$\rightarrow$\texttt{choose}:\textcolor{attr_c}{[attr-A]} $\mid$ \textcolor{attr_c}{[attr-B]}}}
		\\
		\cline{1-1}
		\makecell[l]{which \textcolor{attr_c2}{[attr-type]} is the \textcolor{attr_c}{[attr1]} \textcolor{obj_c}{[obj1]} that \textcolor{attr_c}{[attr2]} \textcolor{obj_c}{[obj2]} \\ is \textcolor{rel_c}{[rel]}, \textcolor{attr_c}{[attr-A]} or \textcolor{attr_c}{[attr-B]}?}
		&
		\\
		\hline
		\makecell[l]{which \textcolor{attr_c2}{[attr-type]} is the \textcolor{attr_c}{[attr]} \textcolor{obj_c}{[obj]}, \textcolor{attr_c}{[attr-A]} or \textcolor{attr_c}{[attr-B]}?}
		&
		\makecell[l]{\texttt{select}:\textcolor{obj_c}{[obj]}$\rightarrow$\texttt{filter}:\textcolor{attr_c}{[attr]}$\rightarrow$\texttt{choose}:\textcolor{attr_c}{[attr-A]} $\mid$ \textcolor{attr_c}{[attr-B]}}   
		\\
		\hline
		\makecell[l]{is the \textcolor{attr_c2}{[attr-type]} of the \textcolor{attr_c}{[attr1]} \textcolor{obj_c}{[obj1]} the same as that\\ of the \textcolor{attr_c}{[attr2]} \textcolor{obj_c}{[obj2]}?}
		&
		\makecell[l]{\texttt{select}:\textcolor{obj_c}{[obj1]}$\rightarrow$\texttt{filter}:\textcolor{attr_c}{[attr1]}$\rightarrow$\texttt{select}:\textcolor{obj_c}{[obj2]}\\$\rightarrow$\texttt{filter}\textcolor{attr_c}{[attr2]}$\rightarrow$\texttt{compare}:$\langle$\textcolor{attr_c2}{[attr-type]}$\rangle$} 
		\\
		\hline
		\makecell[l]{what is the same attributes of \textcolor{attr_c}{[attr1]} \textcolor{obj_c}{[obj1]} and \\ \textcolor{attr_c}{[attr2]} \textcolor{obj_c}{[obj2]}?}
		&
		\makecell[l]{\texttt{select}:\textcolor{obj_c}{[obj1]}$\rightarrow$\texttt{filter}:\textcolor{attr_c}{[attr1]}$\rightarrow$\texttt{select}:\textcolor{obj_c}{[obj2]}\\$\rightarrow$\texttt{filter}\textcolor{attr_c}{[attr2]}$\rightarrow$\texttt{compare}:$\langle$attribute$\rangle$}
		\\
		\hline
		\makecell[l]{is someone \textcolor{act_c}{[act1]} before or after \textcolor{act_c}{[act2]}?}
		&
		\multirow{5}{*}{\makecell[l]{\texttt{select}:\textcolor{act_c}{[act1]}$\rightarrow$\texttt{localize}:\textcolor{act_c}{[act1]}$\rightarrow$\texttt{select}:\textcolor{act_c}{[act2]}\\$\rightarrow$\texttt{localize}:\textcolor{act_c}{[act2]}$\rightarrow$\texttt{compare}:$\langle$time$\rangle$}}
		\\
		\cline{1-1}
		\makecell[l]{is the duration of someone \textcolor{act_c}{[act1]} for longer \\than the duration of \textcolor{act_c}{[act2]}?}
		& 
		\\
		\cline{1-1}
		\makecell[l]{is the duration of someone \textcolor{act_c}{[act1]} for shorter \\than the duration of \textcolor{act_c}{[act2]}?}
		&
		\\
		\hline
		\makecell[l]{is \textcolor{attr_c}{[attr1]} \textcolor{obj_c}{[obj1]} \textcolor{rel_c}{[rel]} \textcolor{attr_c}{[attr2]} \textcolor{obj_c}{[obj2]} and \textcolor{attr_c}{[attr3]} \textcolor{obj_c}{[obj3]}?} 
		&
		\makecell[l]{\texttt{select}:\textcolor{obj_c}{[obj1]}$\rightarrow$\texttt{filter}:\textcolor{attr_c}{[attr1]}$\rightarrow$\texttt{relate}:\textcolor{obj_c}{[obj2]},\textcolor{rel_c}{[rel]}\\$\rightarrow$\texttt{filter}:\textcolor{attr_c}{[attr2]}$\rightarrow$\texttt{and}$\rightarrow$\texttt{relate}:\textcolor{obj_c}{[obj3]},\textcolor{rel_c}{[rel]}\\$\rightarrow$\texttt{filter}:\textcolor{attr_c}{[attr3]}$\rightarrow$\texttt{exist}}
		\\
		\hline
		\makecell[l]{is \textcolor{attr_c}{[attr1]} \textcolor{obj_c}{[obj1]} \textcolor{rel_c}{[rel]} \textcolor{attr_c}{[attr2]} \textcolor{obj_c}{[obj2]} but not \textcolor{attr_c}{[attr3]} \textcolor{obj_c}{[obj3]}?} 
		&
		\makecell[l]{\texttt{select}:\textcolor{obj_c}{[obj1]}$\rightarrow$\texttt{filter}:\textcolor{attr_c}{[attr1]}$\rightarrow$\texttt{relate}:\textcolor{obj_c}{[obj2]},\textcolor{rel_c}{[rel]}\\$\rightarrow$\texttt{filter}:\textcolor{attr_c}{[attr2]}$\rightarrow$\texttt{xor}$\rightarrow$\texttt{relate}:\textcolor{obj_c}{[obj3]},\textcolor{rel_c}{[rel]}\\$\rightarrow$\texttt{filter}:\textcolor{attr_c}{[attr3]}$\rightarrow$\texttt{exist}}  
		\\		
	\end{tabular}
	\vspace{-5pt}
	\caption{Functional programs and their corresponding question templates. Each program consists of a sequence of predefined primary functions. The \texttt{relate} function can support the association of either subject or object. The symbol `\_' means traversing all objects to meet the following constraint.} 
	\label{table:programs}
	\vspace{-10pt}
\end{table*}

\captionsetup[subfigure]{font=normalsize}

\begin{figure*}[h]
	\begin{center}
		\includegraphics[width=0.75\textwidth]{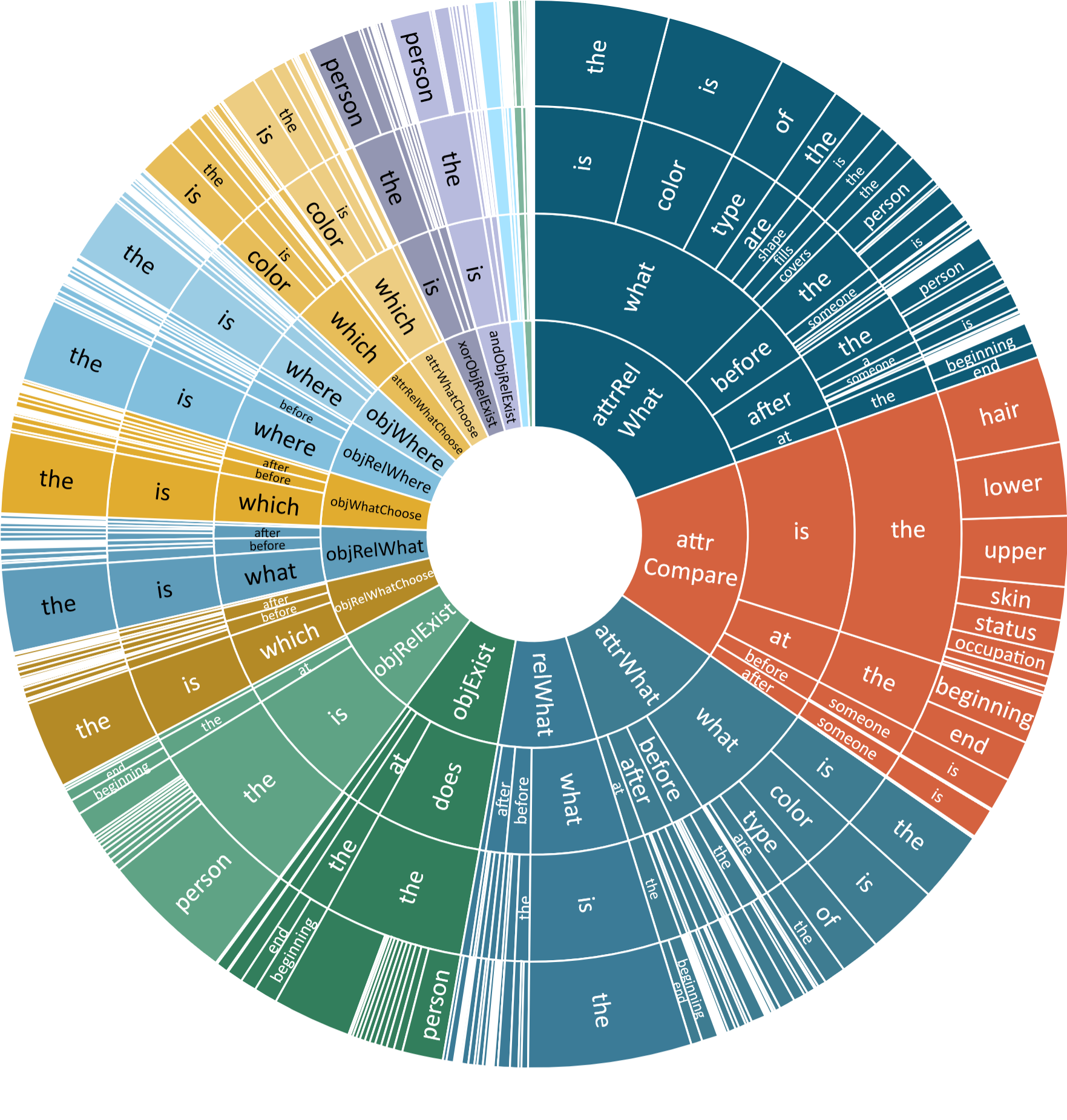}
		\caption{\textbf{Question distribution by their first three words} on the balanced benchmark. The innermost ring refers to the 21 question types. The ordering of the words starts towards the center and radiates outwards. The arc length is proportional to the number of questions containing the word. For the questions with the same structure (query, compare, verify, choose, and logic), we use the background color from the same color scheme (blue, orange, green, yellow, and purple). 
		}
		\vspace{-27pt}
		\label{fig:ques_dist}
	\end{center}
\end{figure*}

\captionsetup[subfigure]{font=normalsize}

\begin{table}
	\centering
	\begin{tabular}{l|ccc}
		& ~HCRN~  & ~ClipBERT~  & ~All-in-one~ \\
		\ChangeRT{1.3pt}
		\texttt{val} & 41.69 & 44.34 & 45.44\\
		\texttt{test} & 41.15 & 43.92 & 44.53\\
		\texttt{test-dev} & 41.18 & 44.00 & 44.57\\		
	\end{tabular}
	\vspace{-5pt}
	\caption{Comparative results of the three models which are trained on the \texttt{train} split and then evaluated on the \texttt{val}, \texttt{test}, and \texttt{test-dev} splits of ANetQA, respectively.} 
	\label{table:val-test}
	\vspace{-5pt}
\end{table}

\begin{table}
	\centering
	\begin{tabular}{l|ccc}
		type & HCRN  & ClipBERT  & All-in-one \\
		\ChangeRT{1.3pt}
		attrRelWhat & 24.06 &	29.03 &	\textbf{29.42} \\
		attrWhat & 21.95 &	26.58 &	\textbf{28.75}  \\
		relWhat &  16.35 &	14.59 &	\textbf{16.94} \\
		objRelWhere &  15.78 &	16.81 &	\textbf{16.21}  \\
		objRelWhat & 19.60 &	19.36 &	\textbf{22.23}   \\
		objWhere &  \textbf{16.34} &	14.25 &	15.39 \\
		objWhat & 39.10 & 39.39 & \textbf{40.11} \\
		objExist & 68.54 &	72.76 &	\textbf{73.20}  \\
		objRelExist &  68.00 &	\textbf{71.85} &	70.92 \\
		actExist & 75.34 &	\textbf{78.04} &	77.85 \\
		objRelWhatChoose & 67.09 &	67.96 &	\textbf{69.13}  \\
		objWhatChoose & 71.51 &	77.63 &	\textbf{77.93}  \\
		attrRelWhatChoose & 56.14 &	64.60 &	\textbf{65.74}  \\
		attrWhatChoose & 57.92 &	65.90 &	\textbf{66.89} \\
		attrCompare &  \textbf{55.66} &	55.60 &	54.42 \\
		attrSame &  56.25 &	\textbf{82.14} &	58.93 \\
		actTime & 67.24 	&\textbf{70.44} &	56.16 \\
		actLongerVerify	 &  50.00 &	50.00 &	\textbf{52.48} \\
		actShorterVerify & 49.79 &	49.79 &	\textbf{50.83} \\
		andObjRelExist &  70.89 &	70.38 &	\textbf{73.97} \\
		xorObjRelExist & 86.50 &	\textbf{89.74} &	87.18 \\		
	\end{tabular}
	\vspace{-5pt}
	\caption{Per-type accuracy of the three models on the \texttt{test} split.} 
	\label{table:per-type-acc}
	\vspace{-5pt}
\end{table}

\clearpage
\newpage
\clearpage

{\small
	\bibliographystyle{ieee_fullname}
	\bibliography{anetqa}
}
\end{document}